\documentclass[single-column]{fairmeta}
\usepackage{titletoc}

\usepackage[T1]{fontenc}    % use 8-bit T1 fonts
\usepackage{hyperref}
\usepackage[most]{tcolorbox}

\usepackage{amsmath}
\usepackage{amssymb}
\usepackage[disable]{todonotes}
\usepackage{listings}
\usepackage{multirow}
\usepackage{makecell}
\usepackage{graphicx}
\usepackage{wrapfig}
\usepackage{float}
\usepackage{tabularx}
\usepackage{textcomp}
\usepackage{stfloats}
\usepackage{url}
\usepackage{verbatim}
\usepackage{titlesec}
\usepackage{tocloft}
\usepackage{adjustbox}
\usepackage{multirow}
\usepackage{pifont}
\usepackage{tikz}
\usepackage{comment}
\usepackage{amsmath,amssymb} % define this before the line numbering.
\usepackage{colortbl}  %
\usepackage{color}
\usepackage{booktabs} 
\usepackage{hyperref}
\usepackage{graphicx}    % 
\usepackage{subcaption} 
\usepackage{multirow} % 
\usepackage{booktabs} % 
\usepackage{subcaption} 
\RequirePackage{xspace}
\makeatletter
\DeclareRobustCommand\onedot{\futurelet\@let@token\@onedot}
\def\@onedot{\ifx\@let@token.\else.\null\fi\xspace}
\usepackage[most]{tcolorbox}
\usepackage{xcolor}
\usepackage{booktabs}
\usepackage{multirow}
\usepackage{makecell}
\usepackage{diagbox}
\usepackage{adjustbox}
\usepackage[table]{xcolor}
\usepackage{xltabular}
\usepackage{array}
\usepackage{threeparttable}

\definecolor{wmgreen}{RGB}{46,125,50}
\definecolor{wmorange}{RGB}{230,145,56}
\definecolor{wmgray}{RGB}{120,120,120}
\usepackage{booktabs}
\usepackage{threeparttable}
\usepackage{pifont}
\usepackage{graphicx}

\definecolor{myblue}{RGB}{233, 241, 249}
\definecolor{mygray}{RGB}{99, 110, 114}
\definecolor{myred}{RGB}{255, 118, 117}
\definecolor{myyellow}{RGB}{255, 234, 167}
\definecolor{mygreen}{RGB}{216, 226, 204}
\definecolor{my_green}{RGB}{51,102,0}
\definecolor{my_red}{RGB}{204, 0, 0}
\renewcommand{\checkmark}{\textcolor{my_green}{\ding{51}}} % ✔
\renewcommand{\texttimes}{\textcolor{my_red}{\ding{55}}} % ✘
\renewcommand{\checkmark}{\textcolor{my_green}{\ding{51}}} % ✔
\renewcommand{\texttimes}{\textcolor{my_red}{\ding{55}}} % ✘

\newcommand{\pmark}{--}

\definecolor{promptframe}{HTML}{5D5D5D}
\definecolor{promptbg}{HTML}{F5F5F5}

\newtcolorbox{promptbox}[1]{
    title=#1,                  
    colframe=promptframe,      
    colback=promptbg,          
    coltitle=white,            
    fonttitle=\large,          
    boxrule=1.5pt,             
    arc=4pt,                   
    left=8pt,                  
    right=8pt,                 
    top=6pt,                   
    bottom=6pt,
    breakable
}
\newtcolorbox{trajbox}[2]{
  enhanced,
  breakable,
  colback=white,
  colframe=#2,
  colbacktitle=#2!90!black,
  boxrule=1pt,
  arc=3pt,
  left=8pt,
  right=8pt,
  top=6pt,
  bottom=6pt,
  title={\textbf{#1}},
  coltitle=white,
  fonttitle=\large,
}

\newcommand{\stepbar}[1]{%
  \vspace{1.2em}
  \noindent\colorbox{gray!15}{%
    \parbox{\dimexpr\linewidth-2\fboxsep}{%
      \vspace{3pt}\textbf{#1}\vspace{3pt}%
    }%
  }\par
  \vspace{0.8em}
}

\newcommand{\tagthink}{\textcolor{purple!70!black}{\texttt{<think>}}}
\newcommand{\stoptagthink}{\textcolor{purple!70!black}{\texttt{</think>}}}
\newcommand{\tagtoolcall}{\textcolor{blue!65!black}{\texttt{<tool\_call>}}}
\newcommand{\stoptagtoolcall}{\textcolor{blue!65!black}{\texttt{</tool\_call>}}}
\newcommand{\tagtoolresponse}{\textcolor{black}{\texttt{<tool\_response>}}}
\newcommand{\stoptagtoolresponse}{\textcolor{black}{\texttt{</tool\_response>}}}
\newcommand{\taganswer}{\textcolor{black}{\texttt{<done>}}}
\newcommand{\stoptaganswer}{\textcolor{black}{\texttt{</done>}}}

\newenvironment{traceblock}
{\par\small\ttfamily\raggedright}
{\par\normalsize}

\title{InterLV-Search: Benchmarking Interleaved Multimodal Agentic Search}
\definecolor{myblue}{RGB}{233, 241, 249}
\definecolor{mygray}{RGB}{99, 110, 114}
\definecolor{myred}{RGB}{255, 118, 117}
\definecolor{myyellow}{RGB}{255, 234, 167}
\definecolor{mygreen}{RGB}{216, 226, 204}
\definecolor{mypurple}{RGB}{162, 155, 254}
\definecolor{mybrown}{RGB}{215, 190, 154}
\definecolor{myorange}{RGB}{255, 220, 190} 
\author[1,3,*]{Bohan~Hou}
\author[2,4,*]{Jiuning~Gu}
\author[3]{Jiayan~Guo}
\author[3]{Ronghao~Dang}
\author[1,3]{Sicong~Leng}
\author[3]{Xin~Li}
\author[4,\dag]{Xuemeng~Song}
\author[1,\dag]{Jianfei~Yang}

\small{
\affiliation[1]{Nanyang~Technological~University}
\affiliation[2]{Shandong~University}
\affiliation[3]{Damo~Academy, Alibaba~Group}
\affiliation[4]{Southern~University~of~Science~and~Technology}
}

\footnotesize{
\contribution[*]{Equal Contribution }
\contribution[\dag]{Corresponding Author}
}

\abstract{
Existing benchmarks for multimodal agentic search evaluate multimodal search and visual browsing, but visual evidence is either confined to the input or treated as an answer endpoint rather than part of an interleaved search trajectory. We introduce \textbf{InterLV-Search}, a benchmark for \textbf{Inter}leaved \textbf{L}anguage-\textbf{V}ision Agentic \textbf{Search}, in which textual and visual evidence is repeatedly used to condition later search. It contains 2,061 examples across three levels:  active visual evidence seeking, controlled offline interleaved multimodal search, and open-web interleaved multimodal search. Beyond existing benchmarks, it also includes multimodal multi-branch samples that involve comparison between multiple entities during the evidence search.
We construct Level~1 and Level~2 with automated pipelines and Level~3 with a machine-led, human-supervised open-web pipeline. We further provide \textbf{InterLV-Agent} for standardized tool use, trajectory logging, and evaluation. Experiments on proprietary and open-source multimodal agents show that current systems remain far from solving interleaved multimodal search, with the best model below 50\% overall accuracy, highlighting challenges in visual evidence seeking, search control, and multimodal evidence integration.
 We release the benchmark data and evaluation code at \url{https://github.com/hbhalpha/InterLV-Search-Bench}.
}

\begin{document}

\maketitle
\section{Introduction}
Recent advances in large language models (LLMs) have spurred the development of multimodal large language models (MLLMs), enabling strong multimodal understanding via large-scale pretraining. These models are highly effective when all required context is contained in the multimodal input~\citep{guo2025deepseek}, supporting reliable in-context multimodal reasoning. However, many real-world tasks, such as question answering~\citep{marino2019ok,chang2022webqa} and deep research~\citep{huang2026mmdeepresearch,narayan2025deepmmsearch}, are open-world and cannot be resolved solely from the provided input, as necessary evidence often lies beyond the observed context and requires external information access. This has motivated growing interest in multimodal agentic search~\citep{wu2025mmsearch,chng2025sensenova}, where models actively plan, invoke tools~\citep{yao2023react}, retrieve and browse webpages~\citep{koh2024visualwebarena} and images, inspect visual evidence, and synthesize information across heterogeneous sources.
%The web, with its broad coverage, dynamic updates, and heterogeneous multimodal content, serves as a particularly rich source for such needs. 
%Such paradigms have demonstrated promising success in open-world information seeking tasks, including question answering~\citep{} and deep-research-style systems~\citep{huang2026mmdeepresearch,narayan2025deepmmsearch}, highlighting their effectiveness in complex real-world scenarios.

%A key capability of multimodal agentic search system is to actively search over external textual and visual evidence, rather than only reason over evidence provided in the input. 

%many multimodal search benchmarks~\citep{wu2025mmsearch,tao2025mmsearch,zeng2026vision,geng2025webwatcher} provide samples introduce visual inputs such as images, crops, screenshots, or \textcolor{red}{other visual contexts}, thereby testing whether models can reason over visual evidence in search-like settings. However, such visual evidence is largely pre-specified by the benchmark rather than actively sought by the agent during search. 

\begin{figure*}[t]
    \centering
    \includegraphics[width=\linewidth]{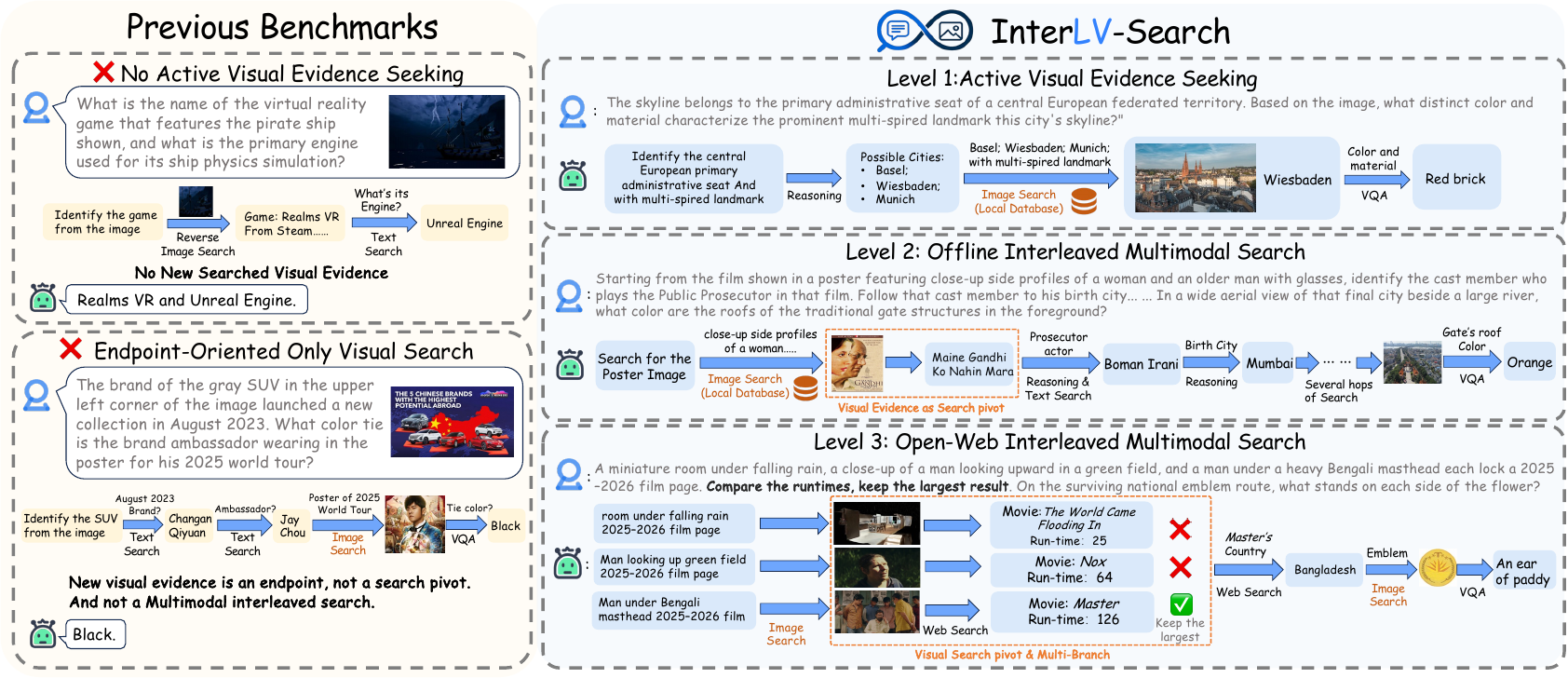}
    \caption{
    Comparison of our benchmark with prior benchmarks.
    }
    \vspace{-1em}
    \label{fig:teaser_compare}
\end{figure*}

As illustrated in the upper-left panel of Fig.~\ref{fig:teaser_compare}, early benchmarks~\citep{wu2025mmsearch,jiang2024mmsearch,zeng2026vision,geng2026webwatcher} for multimodal agentic search largely focus on evaluating textual evidence acquisition, with visual information restricted to the initial user input in various forms, e.g., images, cropped regions, screenshots, and other visual contexts. To incorporate visual information during evidence retrieval, recent visual browsing benchmarks, including VisBrowse~\citep{visbrowse} and BrowseComp-$V^3$~\citep{zhang2026browsecomp}, further require models to explicitly locate relevant visual entities or images. However, as shown in the lower-left panel of Fig.~\ref{fig:teaser_compare}, these benchmarks still treat retrieved visual evidence as an answer-bearing endpoint: once a relevant image or visual entity is found, it is primarily used to answer a local visual question and support final answer derivation. %This formulation overlooks an alternative but critical role of visual evidence in the search trajectory, namely its \emph{search-controlling} function. 
This formulation overlooks an alternative but critical role of visual evidence in the search trajectory: visual evidence can be \emph{search-controlling}, determining what the agent should search for next. In realistic information seeking, visual observations often contain fine-grained cues---such as logos, inscriptions, persons, or spatial relations~\citep{tao2026mmsearchplus}---that disambiguate the current state and reveal the next search target, including the next query, entity, webpage, tool invocation, or branching decision, as illustrated in the right panel of Fig.~\ref{fig:teaser_compare}.

%Rather than evaluating whether an agent can merely retrieve visual evidence, we assess whether intermediate textual and visual observations can drive subsequent retrieval. A search trajectory is interleaved when evidence from one modality conditions later retrieval or inspection in another modality, creating cross-modal dependencies throughout the search process.

%To address these limitations, we aim to  evaluate whether agents can leverage visual evidence to control long-horizon search trajectories, including multi-hop visual retrieval, repeated cross-modal transitions, and branch-conditioned evidence integration.  However, existing benchmarks fail to capture this evaluation dimension. 

%As illustrated in Fig.~\ref{fig:teaser_compare}, endpoint-oriented visual browsing retrieves an image and then answers a local visual question. In contrast, interleaved multimodal search treats visual evidence as a search pivot: a logo, inscription, person, emblem component, spatial relation, or other visual cue may determine the next query, entity, page, tool call, or branch decision. This shifts the evaluation from answering after visual evidence is found to searching through visual evidence. The comparison between our benchmark and prior work is summarized in Table~\ref{tab:benchmark_comparison}.

\begin{table*}[t]
%\centering
\scriptsize
\setlength{\tabcolsep}{2.2pt}
\renewcommand{\arraystretch}{1.12}
\caption{{Benchmark Comparison.} Textual/visual multi-hop requires $\ge$2 retrieval hops over textual/visual evidence.  Auto and Semi-auto denote fully automated and semi-automated construction, respectively. %\texttimes\ is assigned if $\ge$90\% of samples in the benchmark do not require the capability.
For prior benchmarks, we manually inspect all their samples and assign \texttimes\ only if at least 90\% of samples do not require the capability. 
}
\begin{tabular*}{\linewidth}{@{\extracolsep{\fill}}l c c c c c c c c}
\toprule
Benchmark & Samples & 
\makecell{ Active Visual\\ Evidence Seeking} &
\makecell{Textual\\Multi-hop} &
\makecell{Visual\\Multi-hop} &
\makecell{Recurrent V--T \\Interleaving} &
\makecell{ Role  of Retrieved\\  Visual Evidence} &
\makecell{Multi-\\Branch} &
Construction \\
\midrule
FVQA-Test & 1800 & \texttimes & \texttimes & \texttimes & \texttimes & - & \texttimes & Semi-auto+Manual \\
MMSearch & 300 & \texttimes & \checkmark & \texttimes & \texttimes & - & \texttimes & Manual \\
MMSearch-Plus & 311 & \texttimes & \checkmark & \texttimes & \texttimes & -  & \texttimes & Semi-auto \\
BrowseComp-VL & 399 & \texttimes & \checkmark & \texttimes & \texttimes & - & \texttimes & Semi-auto \\
VDR-Bench & 2,000 & \texttimes & \checkmark & \texttimes & \texttimes & - & \texttimes & Semi-auto \\
BrowseComp-$V^3$ & 300 & \checkmark & \checkmark & \texttimes & \texttimes & Endpoint & \texttimes & Manual \\
VisBrowse & 169 & \checkmark & \checkmark & \texttimes & \texttimes & Endpoint & \texttimes & Manual \\
\midrule
\textbf{InterLV-Search} & \textbf{2,061} & \checkmark & \checkmark & \checkmark & \checkmark & \textbf{Pivot+Endpoint} & \checkmark & Auto+Semi-auto \\
\bottomrule
\end{tabular*}
\label{tab:benchmark_comparison}
\vspace{-1.5em}
\end{table*}

%A visual step is search-controlling if information extracted from visual evidence determines a subsequent retrieval target, such as the next query, entity, page, tool call,  branch or done decision. 

%In contrast, to fully leverage visual evidence and address real-world, complex user queries, systems should enable vision–text interleaved evidence retrieval. In this paradigm, not only is intermediate text reasoning used to retrieve visual evidence, but intermediate visual evidence containing rich fine-grained cues (e.g., a logo, inscription, person, emblem component, or spatial relation) also informs subsequent retrieval targets, such as the next query, entity, webpage, tool call, or branching decision, as illustrated in the right panel of Figure~\ref{fig:teaser_compare}. However, existing benchmarks do not support the evaluation of this capability.

Motivated by this gap, we formulate \emph{interleaved multimodal search} as the target capability of our benchmark, emphasizing that intermediate visual evidence should serve not only as a source for question answering but also as a signal that guides subsequent retrieval decisions. In this setting, an agentic search system must dynamically switch between visual and textual evidence acquisition, where evidence from one modality determines subsequent retrieval actions in the other. Specifically, we require recurrent vision--text interleaving, such that after merging consecutive same-modality steps, each trajectory contains multiple visual segments with textual search or reasoning in between~\citep{du2025easy}, and later retrieval is conditioned on earlier evidence.

To evaluate this capability, we introduce \textbf{InterLV-Search}, a three-level benchmark for \textbf{Inter}leaved \textbf{L}anguage-\textbf{V}ision Agentic \textbf{Search}. InterLV-Search decomposes interleaved multimodal search into progressively challenging settings: active visual evidence seeking (Level~1), offline interleaved search (Level~2), and in-the-wild open-web search (Level~3). Level~1 evaluates active visual evidence seeking from textual information needs, the primitive ability to use vision signals in agentic search. Level~2 tests whether agents can perform multi-hop interleaved evidence search in a controlled offline environment~\citep{deng2026deepimagesearchbenchmarkingmultimodalagents}, avoiding confounders such as ranking instability, page variation, and non-unique evidence sources in real-world environments. Level~3 evaluates the same mechanism in an in-the-wild open-web setting~\citep{zhou2024webarena,koh2024visualwebarena}, where agents face noisy, dynamic webpages, images, and search results. %Unlike the controlled setting, open-web search may also branch into multiple routes. 
To meet diverse practical demands, Level~3 includes both standard single-chain examples and multi-branch examples that involve comparisons among multiple entities during evidence search, where the agent must explore multiple branches, gather textual or visual evidence, and proceed along a selected branch.
%, where agents must search along multiple parallel routes, collect textual or visual evidence from each route, compare and reason over attributes of the evidence (e.g., year or size), and continue searching from the selected branch. 
This enables InterLV-Search to evaluate non-linear search control beyond prior single-chain multimodal search benchmarks. %Notably, this level is designed to support both linear and branching evidence search paradigms. %While linear search, where agents perform sequential search steps conditioned on the current observation, has been widely adopted in existing multimodal agentic search studies, branching-based search, which allows more exploratory behavior with multiple search trajectories pursued in parallel, remains unexplored. %Overall, our staged design isolates the core evidence-to-query mechanism before testing its robustness under realistic web interaction. 

To scale InterLV-Search, we develop fully automatic MLLM-driven pipelines that involve internal filtering and verification for Level~1 and Level~2 construction, leveraging high-quality multimodal entity data and knowledge-graph chains in MMKG-W~\citep{zhang2025mmkg}, a Wikimedia-based multimodal knowledge graph containing around 15K entities. Level~3 adopts a machine-led, human-supervised process, where web-capable agents generate open-world QA pairs requiring interleaved multimodal evidence search, and expert annotators provide feedback and corrections. %To instantiate InterLV-Search at scale, we develop automated and semi-automated construction pipelines. Level~1 and Level~2 are generated through automatic pipelines over high-quality multimodal entity data and knowledge-graph chains, followed by filtering and verification. Level~3 uses a machine-led, human-supervised process, where web-capable agents generate open-web multimodal interleaved search questions and answers, and expert annotators provide feedback and corrections.
Together, these pipelines produce 2,061 examples across three levels. As shown in Table~\ref{tab:benchmark_comparison}, InterLV-Search is, to the best of our knowledge, the first benchmark to jointly cover text-to-visual search, visual multi-hop retrieval, recurrent vision--text interleaving, and multimodal multi-branch search.

To standardize evaluation on InterLV-Search, we implement \textbf{InterLV-Agent}, a reference framework for unified tool use, trajectory logging, and model comparison. Using this framework, we evaluate both proprietary and open-source multimodal agents. Experiments show that current models still struggle with interleaved multimodal search and evidence integration: even with tool use, the best model remains below 50\% overall accuracy.

Our main contributions are summarized as follows:
\begin{itemize}
\item \textbf{InterLV-Search Benchmark.}
It contains 2,061 examples across three progressively challenging levels, enabling the evaluation of agentic systems in visual evidence seeking, as well as offline and open-web interleaved multimodal evidence search.  %It evaluates whether retrieved visual evidence can serve not only as an endpoint, but also as a search pivot that drives subsequent retrieval.

    \item \textbf{Scalable data construction pipelines.}
    We build automated pipelines for Level~1 and Level~2, and a machine-led, human-supervised semi-automated pipeline for Level~3, enabling scalable construction of high-quality  interleaved multimodal search data. We will release the construction pipelines upon publication.  %These pipelines generate, filter, and verify multimodal search chains. % with controlled evidence paths, reducing leakage, ambiguity, and shortcut solutions.

    \item \textbf{Comprehensive evaluation and analysis.}
We evaluate proprietary and open-source multimodal agents on InterLV-Search and provide detailed analyses, revealing that current models still struggle with interleaved multimodal search.
\end{itemize}

\section{InterLV-Search Benchmark}

%\subsection{Overview}

To construct a comprehensive benchmark for interleaved multimodal search, we organize InterLV-Search into three progressively challenging levels: visual evidence seeking (Level~1), controlled interleaved search (Level~2), and in-the-wild open-web search (Level~3). This design mirrors the capability progression required of multimodal search agents: an agent must first acquire missing visual evidence, then integrate such evidence into multi-hop evidence-to-query transitions, and ultimately execute the same search paradigm in the open-web setting.

We adopt different construction strategies according to the controllability of each level. Level~1 and Level~2 are constructed with fully automated pipelines, where we use Gemini-3.1-Pro~\citep{googledeepmind2026gemini31pro} as the generator, composer, and verifier for producing search needs, visual queries, interleaved chains, and quality judgments. Level~3 involves real webpages and noisier evidence sources, so we adopt a semi-automated pipeline: GPT-5.4-Thinking~\citep{openai2026gpt54thinking} serves as a web-search-capable generation agent~\citep{du2026deepresearch} for automated candidate construction, while PhD-level human participants provide human-in-the-loop verification and refinement to ensure evidence validity, answerability, and high-quality search chains.
%In this section, we first describe the construction pipeline for each level, and then summarize the benchmark statistics.

%\subsection{Benchmark Construction Pipeline}
%To enable scalable benchmark construction, we use fully automated construction and filtering pipelines for Level~1 and Level~2, allowing us to synthesize large-scale evaluation samples at low cost. For Level~3, where open-web sources are noisier and more variable, we rely primarily on automated candidate generation but incorporate human-in-the-loop verification and refinement to ensure evidence validity, answerability, and search-chain quality.

\subsection{Level 1: Active Visual Evidence Seeking}
Level~1 evaluates a system’s ability to seek visual evidence from textual information needs, a fundamental capability for interleaved search. % without reliably retrieving and inspecting visual evidence, an agent cannot leverage visual cues as intermediate pivots in subsequent multi-hop reasoning. 
We formulate this level as a \emph{Search-to-VQA} task~\citep{luo2021weakly,hong2026knowledgebased} (Fig.~\ref{fig:teaser_compare}), where each question encodes a fine-grained visual query about an implicitly specified target entity. To answer it, the agent must first infer and retrieve the hidden entity from the query, and then inspect the corresponding image. The final answer is not the entity name, but a concise image-grounded attribute, such as color, object, count, material, pattern, or spatial relation.

\begin{figure*}[t]
    \centering
    \includegraphics[width=0.95\linewidth]{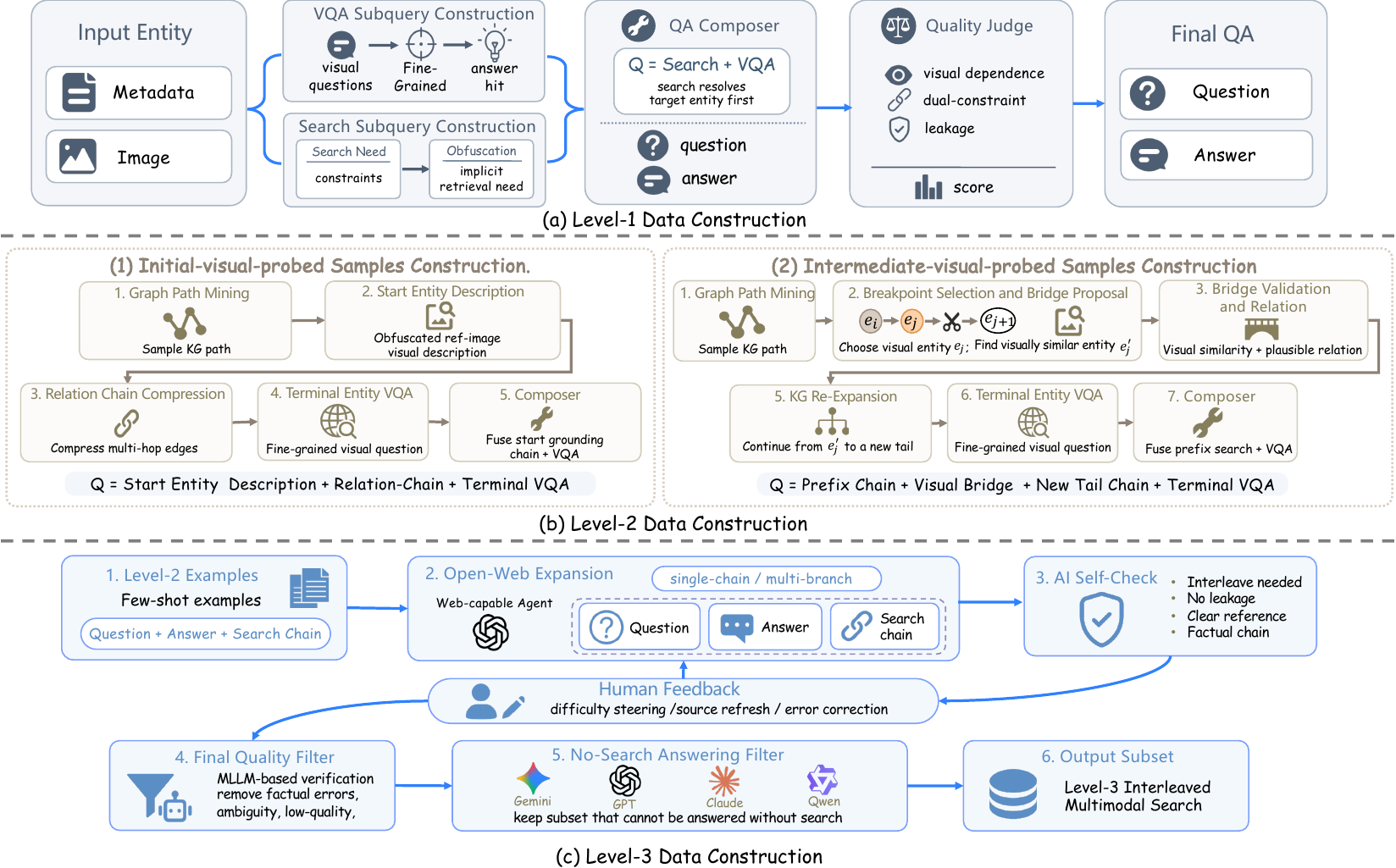}
    \caption{
    Data Construction Pipeline of InterLV-Search Benchmark. %   \textbf{(a) Construction of Level-1 Data. }     \textbf{(b) Construction of Level-2 Data }    \textbf{(c) Construction of Level-3 Data }
    }
    \label{fig:constrcution}
        \vspace{-1em}
\end{figure*}

\textbf{Data Source.}
We construct Level~1 from MMKG-W~\citep{zhang2025mmkg}, a Wikimedia-based multimodal knowledge graph containing approximately 15K entities. Each entity is associated with a canonical Wikidata item (i.e., a unique entity identifier), an image, and textual metadata such as a description field and a ``what is it'' field. This source is well-suited for Level~1: the metadata provides searchable semantic anchors, while the paired image serves as grounded visual evidence for answering the final query.

\textbf{Data Construction Pipeline.}
Each Search-to-VQA instance can be decomposed into two components: an implicit target search subquery and a corresponding VQA subquery. Accordingly, as shown in Fig.~\ref{fig:constrcution}(a), our pipeline first constructs these two components for a given entity from MMKG-W, and then composes them into candidate question–answer pairs. We further apply quality filtering to remove low-quality pairs. Since the answer to each instance is directly determined by the VQA subquery, we first instruct an MLLM to construct the VQA component for a given entity, i.e., a fine-grained question–answer pair whose answer (i.e., an image-grounded attribute) cannot be inferred without inspecting the image~\citep{goyal2017making}. Next, we prompt the MLLM to generate an implicit target-search subquery based on the entity’s metadata and corresponding image, while avoiding explicitly naming the entity~\citep{faggioli2024query} or revealing the final visual answer. Finally, rather than simply concatenating the two subqueries, which would make the question unnatural and overexpose the search intent, we use the MLLM to compose them into a single natural question.
%but is rewritten to avoid directly naming the entity or revealing the final visual answer.

%This yields a coupled Search-to-VQA task rather than two loosely connected subtasks.
%Although the final question presents the search need before the visual query, the construction pipeline first identifies the visual query to ensure strong image dependence. 
 %This step identifies a visual attribute, object, count, pattern, material, or spatial relation that will become the final answer target. 

\textbf{Post-processing and Filtering.}
This stage checks whether the composed question truly requires both search and visual inspection. We remove samples that collapse into standalone search or VQA, commonsense guessing, or metadata lookup. We also discard cases with entity or answer leakage, ambiguous targets, or entity-label answers rather than image-grounded attributes. A final judge is used to score each candidate for visual dependence, search specificity, answerability, leakage control, image groundedness, and Search-to-VQA coupling. To validate MLLM-based judging, we manually inspected a subset of judgments  from multiple judge models and found high human agreement; the same validation is applied to subsequent MLLM-based filtering stages.

%Level~2 evaluates controlled interleaved search in an offline environment. This level preserves the core search structure of interleaved multimodal search, while reducing live-web retrieval noise such as search-ranking instability, webpage updates, inaccessible sources, duplicated or conflicting evidence, and non-unique visual results. By controlling the retrieval environment, Level~2 allows us to test whether agents can follow the intended evidence-to-query transitions without confounding failures from open-web uncertainty. 

\subsection{Level 2: Controlled Offline Interleaved Search}
%Level~2 evaluates a more complex form of multimodal search built upon the Level~1 primitive. 
While Level~1 tests whether an agent can actively acquire missing visual evidence, Level~2 examines whether such visual evidence can be used as intermediate pivots in a multi-hop search process, especially in a controlled offline environment. We require each instance to involve at least two rounds of visual evidence retrieval. Since the final fine-grained VQA counts as one round, the agent must first ground a visual clue, convert it into the next retrieval target, and finally ground the terminal image to answer the question. Specifically, we construct Level~2 examples in two complementary forms: initial-visual-probed and intermediate-visual-probed samples, by explicitly introducing visual evidence probes seeking at the beginning or an intermediate stage of the reasoning chain. %For both types, an LLM is used to convert structured paths into implicit search queries, integrate them with terminal VQA sub-queries, and verify that the resulting questions require the intended interleaved search process.

%As a mid-level setting, it focuses on controlled interleaved search in an offline environment, thereby isolating the evaluation from confounding factors (e.g., ranking instability, page variation, and non-unique evidence sources) introduced by open-web uncertainty. Specifically, 
%By constructing these chains offline from a multimodal knowledge graph, Level~2 preserves the intended evidence-to-query transitions while avoiding live-web noise such as ranking instability, page variation, and non-unique evidence sources.

\textbf{Data Source and Chain Mining.}
Level~2 reuses MMKG-W and, building upon Level~1, additionally leverages entity-relation annotations in the knowledge graph (KG) to construct instances for interleaved multimodal search. MMKG-W provides graph edges that connect entities through semantic relations, enabling the extraction of verifiable multi-hop entity paths. Semantic relations between multimodal entities along these paths can inherently act as hidden evidence paths that support the construction of our two types of instances. During path mining, we additionally require the start and terminal entities to be non-adjacent in the KG, reducing shortcut paths for subsequent construction.
%For both types, an LLM is used to convert mined structured paths into implicit search queries, integrate them with terminal VQA sub-queries, and verify that the resulting questions require the intended interleaved search process.

 % form a controllable basis for constructing interleaved search chains that integrate symbolic entity relations with visual search pivots.

%The interleaved search designed in this work requires at least two rounds of visual evidence retrieval. Since the final fine-grained VQA accounts for one retrieval step, we explicitly introduce two additional retrieval stages at the beginning and in the intermediate step. Specifically,

% They differ in where explicit visual search is injected into the chain. 

%Prefix-visual-grounded sample construction introduces visual evidence seeking at the starting node of the path, while intermediate-visual-grounded sample construction forces a visual bridge at an intermediate breakpoint. 
%
 
\textbf{Initial-visual-probed Samples Construction.}
This module explicitly injects visual evidence probing at the beginning of the reasoning chain, requiring the agent to establish the initial search state through visual grounding. Specifically, drawing inspiration from composed image retrieval (CIR)~\citep{song2025comprehensive,hou2025fire}, which retrieves a target image by composing a reference image with textual modification constraints, and given a multi-hop knowledge graph path
$P: e_0 \xrightarrow{r_1} e_1 \xrightarrow{r_2} \cdots \xrightarrow{r_k} e_k$, we regard $e_0$ as the reference entity, while the relations ${r_i}$ together with the textual descriptions of intermediate entities serve as compositional modifications that guide the transition from $e_0$ to $e_k$. To inject initial visual probing, $e_0$ is not directly provided; instead, we use an MLLM to generate an implicit entity query that summarizes the salient visual and semantic cues of this entity. Ultimately, we employ an MLLM to compress and obfuscate the multi-hop path with the initial entity replaced by an implicit entity query~\citep{hou2025fire} into the final natural-language query that implicitly requires interleaved multimodal evidence search without exposing any triple $(e_i, r_i, e_{i+1})$. %The terminal entity $e_k$ is then paired with a fine-grained VQA query over its image, following the same formulation as Level~1. 
\textbf{Intermediate-visual-probed Samples Construction.}
This module generates intermediate-visual-probed samples that require middle-stage visual grounding within the reasoning chain. As shown in Fig.~\ref{fig:constrcution}(b), the construction proceeds in three stages. \textit{1) Visual Breakpoint Selection and Bridge Proposal.} Given a candidate KG path from MMKG-W, e.g., $e_0 \rightarrow e_1 \rightarrow e_2 \rightarrow \cdots \rightarrow e_k$, we first employ an MLLM to select an intermediate entity $e_j$ that exhibits distinctive visual characteristics and serves as a visual breakpoint for subsequent reasoning. The original downstream continuation of the path (i.e., $e_j \rightarrow e_{j+1} \rightarrow \cdots \rightarrow e_k$) is then discarded. Instead, we re-anchor the reasoning process by introducing a bridge entity $e_j'$, retrieved from MMKG-W conditioned on $e_j$, and required to be highly visually similar to $e_j$. \textit{2) Bridge Entity Validation and Bridge Relation Annotation.} To ensure the validity of the bridge entity, a secondary MLLM-based validator verifies that each candidate bridge entity is not only visually similar to $e_j$ but also supported by a plausible semantic relation that justifies transitioning from $e_j$ to $e_j'$. For accepted candidates, the MLLM further annotates the relation between $e_j$ and $e_j'$. This transition inherently requires the agentic search system to first perform text-to-image retrieval conditioned on the image of $e_j$, and then conduct image-to-image retrieval to obtain $e_j'$. \textit{3) KG Re-expansion and Final Question Generation.} Starting from $e_j'$, we resume KG traversal to construct a new tail path (e.g., via multi-hop neighbors), redirecting the reasoning chain after an explicit visual retrieval step to enable subsequent textual multi-hop search. Finally, we construct a Level~1-style fine-grained VQA subquery for the terminal entity $e_k$ and integrate it with the hidden search chain to form the final natural-language question with MLLM rewriting~\citep{ye2023enhancing}.

\textbf{Post-processing and Filtering.} 
%For both construction patterns, the resulting chain is used as a hidden evidence path rather than a question template. After the terminal VQA target is attached, we use a question writer to compose the relation chain, visual pivots, and final visual query into a single coherent question. 
We apply attacker-style checks and judge-based filtering to remove samples that can be solved via direct guessing or lightweight search, leak the target entity or final answer, contain ambiguous visual bridges, or fail to properly couple the search path with the terminal visual question. For intermediate-visual-probed samples, we additionally enforce bridge plausibility, bridge uniqueness, and relation validity before accepting each generated instance.

\vspace{-1em}
\subsection{Level 3: Open-Web Interleaved Multimodal Search}
%Level~3 evaluates open-web interleaved search, where agents must execute the same search mechanism under realistic conditions, handling noisy, dynamic, and non-unique webpages, search results, and evidence sources.

Level~3 evaluates the same interleaved multimodal search capability as Level~2, but in a real open-web setting rather than a controlled offline graph. In this setting, agents operate over noisy webpages, search results, and heterogeneous online sources, where evidence is dynamic, ambiguous, and not globally consistent. The large and heterogeneous open-web source space provides rich and diverse information, which naturally enables questions involving multiple comparable entities. This, in turn, supports both recurrent single-chain search and multi-branch exploration, where different entity-specific evidence sources must be collected and compared.
%The large and heterogeneous open-web source space naturally supports both recurrent single-chain search and multi-branch construction. 
Accordingly, beyond existing benchmarks that focus on single-chain search, we further consider multi-branch interleaved search, where multiple reasoning routes are explored in parallel and selectively continued based on evidence.

%Accordingly, beyond existing benchmarks that only evaluate single-chain multimodal search, we further consider instances that support multi-branch open-web interleaved search, where multiple candidate reasoning routes can be explored in parallel, compared, and selectively continued based on evidence. This setting provides a more stringent evaluation of search control and information integration, requiring agents to maintain multiple active branches, extract comparable evidence across routes, and make decisions based on cross-branch comparison.

%Level~3 evaluates the same interleaved multimodal search capability as Level~2, but in the real open web rather than a controlled offline graph. In this setting, agents must operate over noisy webpages, search results, and heterogeneous online sources, where evidence may be dynamic, non-unique, or visually ambiguous. We include both single-chain open-web interleaved search, where visual and textual pivots guide one recurrent search trajectory, and, to the best of our knowledge, first introduce multi-branch open-web interleaved search, where multiple routes are explored in parallel, compared, and continued from the selected branch. This multi-branch setting more directly evaluates complex search control and information integration, requiring agents to maintain branch states, extract comparable evidence across routes, and use the comparison result to decide how the search should proceed.

\textbf{Data Construction Pipeline.} %Unlike fully manual curation in existing benchmarks, our pipeline delegates source discovery and chain construction to a web-search-capable agent, while humans provide task specification, factual correction, and quality supervision. 
As shown in Fig.~\ref{fig:constrcution}, unlike fully manual curation in existing benchmarks, we construct Level~3 via a semi-automated, human-in-the-loop open-web generation pipeline. 
Specifically, we provide GPT-5.4-Thinking with an explicit task definition (i.e., single-chain or multi-branch) and Level~2 exemplars that illustrate the desired question-answer format and interleaved search-chain structure. Conditioned on this input, GPT-5.4-Thinking generates seed questions, performs web search to retrieve relevant sources, and produces candidate questions, answers, and evidence chains. In particular, for single-chain tasks, it is instructed to construct a linear evidence-to-query trajectory in which intermediate textual or visual evidence progressively guides subsequent retrieval steps. For multi-branch tasks, it is instructed to explore multiple parallel reasoning routes, collect comparable evidence across branches, and formulate a comparison query to guide which branch should be further expanded.
%For multi-branch tasks, it is instructed to explore multiple parallel reasoning routes, collect comparable evidence across branches, and proposes a comparison query that determines which branch should be further expanded.
%ultimately enabling selective continuation based on cross-branch evidence comparison.

Then an AI self-check stage verifies whether each candidate question requires interleaved open-web search, satisfies the specified single-chain or multi-branch constraint, avoids entity or answer leakage, and follows a factual evidence chain. Candidates failing these checks are revised or discarded before final filtering. Meanwhile, PhD-level human annotators review intermediate outputs and provide high-level feedback when the generated chain is insufficiently interleaved, contains spurious multi-hop steps, has weak visual pivots, relies on unreliable sources, exhibits ambiguous constraints, or includes factual inconsistencies. When necessary, they guide GPT-5.4-Thinking to  strengthen visual pivots, revise source selection, or reconstruct the evidence chain.

\textbf{Post-processing and Filtering.}
After candidate generation, we first apply the GPT-5.4-Thinking as quality filter to remove samples with factual errors, ambiguous references, low-quality evidence, answer leakage, broken evidence chains, and unstable webpage sources to reduce answer drift over time. We then apply a no-search answering filter to reduce shortcuts from memorized knowledge. Specifically, we ask multiple strong models, including Gemini~\citep{comanici2025gemini}, GPT~\citep{openai2026gpt54thinking}, Claude~\citep{anthropic2026claudesonnet46}, and Qwen~\citep{qwen2026qwen36plus}, to answer each candidate without  web search. We use their responses to estimate how likely each question can be solved from parametric knowledge~\citep{roberts2020much} alone.

To select the final subset, we formulate a subset-selection problem and solve it with a CP-SAT optimizer~\citep{perron2023cpsatlp}. The selected subset is required to satisfy three criteria: low average no-search accuracy, balanced difficulty across different model families, and a minimum retention size. This prevents the final benchmark from containing questions that are easy to answer without search, while also avoiding a subset that only exploits the weakness of a single model. % From roughly 2,300 automatically collected candidates, this filtering process retains 861 Level~3 examples.
\vspace{-1em}
\subsection{Benchmark Statistics}

\begin{figure*}[t]
    \centering
    \begin{minipage}[c]{0.62\textwidth}
        \centering
        \includegraphics[width=\linewidth]{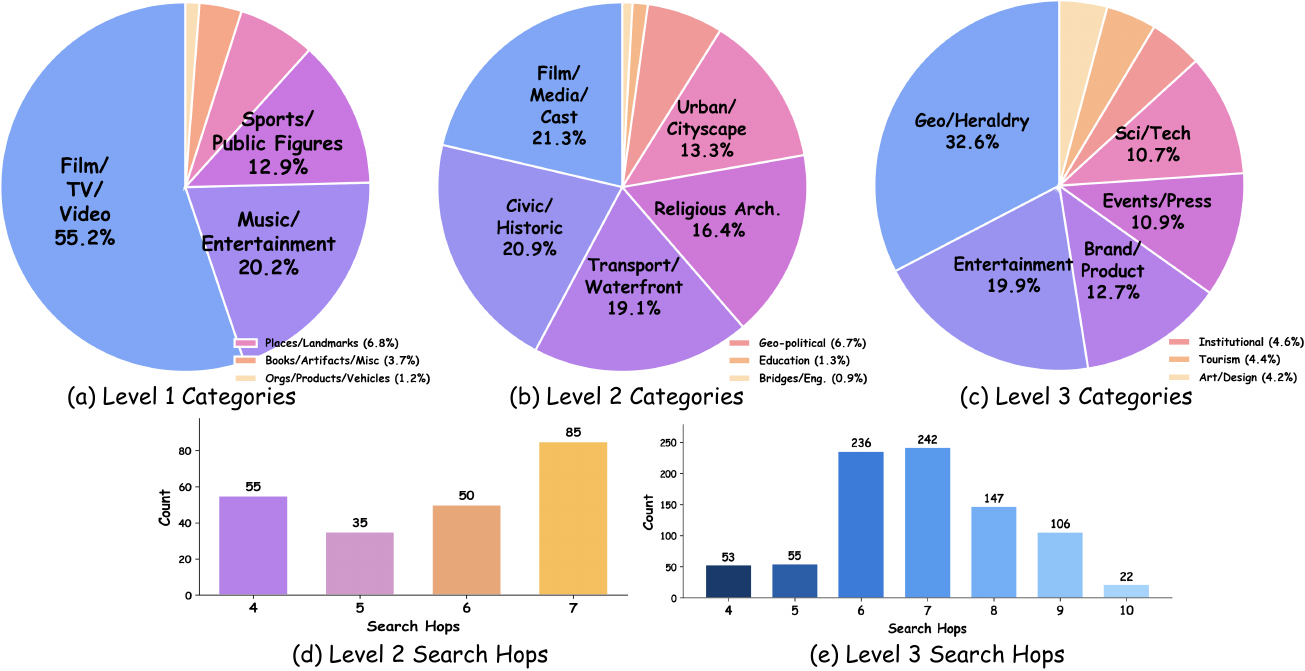}
    \end{minipage}
    \hspace{0.025\textwidth}
    \begin{minipage}[c]{0.3\textwidth}
        \centering
        \small
        \resizebox{\linewidth}{!}{%
        \begin{tabular}{llr}
        \toprule
        \textbf{Type} & \textbf{Statistic} & \textbf{Number} \\
        \midrule
        \multirow{4}{*}{Level 1}
        & Total questions & 975 \\
        & Offline retrieval pool size & 14{,}943 \\
        & Average question length & 42.57 \\
        & Average answer length & 3.59 \\
        \midrule
        \multirow{6}{*}{Level 2}
        & Total questions & 225 \\
        & Offline retrieval pool size & 14{,}943 \\
        & Average question length & 67.92 \\
        & Average answer length & 2.93 \\
        & Initial-visual-probed samples & 125 \\
        &Intermediate-visual-probed samples & 100 \\
        \midrule
        \multirow{5}{*}{Level 3}
        & Total questions & 861 \\
        & Average question length & 49.02 \\
        & Average answer length & 2.50 \\
        & Single-chain questions & 521 \\
        & Multi-branch questions & 340 \\
        \midrule
        \multirow{4}{*}{Overall}
        & Total questions & 2{,}061 \\
        & Offline retrieval pool size & 14{,}943 \\
        & Average question length & 48.03 \\
        & Average answer length & 3.06 \\
        \bottomrule
        \end{tabular}%
        }
    \end{minipage}
    \caption{
    Statistics of InterLV-Search. 
    Left: category and search-hop distributions across benchmark levels. % Level~1 covers diverse searchable visual entities, while Level~2 and Level~3 contain non-trivial multi-hop search chains. 
    Right: overall benchmark statistics. %, including dataset size, offline retrieval pool size, average question and answer length, and the question type distribution for Level~3.
    }
    \label{fig:benchmark_statistics}
    \vspace{-1em}

\end{figure*}
As shown in Fig.~\ref{fig:benchmark_statistics}, InterLV-Search contains 2,061 examples: 975 for Level~1, 225 for Level~2, and 861 for Level~3. Across the three levels, it covers diverse visual and open-web domains, including entertainment, public figures, places, organizations, products, geographic symbols, events, science and technology, tourism, and art. Level~2 and 3 contain multi-hop chains with average estimated lengths of 6.0 and 6.9 hops. Level~3 further includes 340 multi-branch examples (39.5\%), enabling evaluation of parallel-route search and evidence-based branch selection.

\section{InterLV-Agent}

To standardize evaluation on InterLV-Search, we implement \textbf{InterLV-Agent}, a reference framework for interleaved language--vision search. It follows a reason-act-observe loop and provides unified tool use, trajectory logging, and model comparison. The supported tools include image search, reverse image search, web search, webpage browsing, image cropping, and code execution; for Level~1 and Level~2, an offline multimodal retriever enables controlled retrieval over the benchmark corpus. For Levels 2 and 3, InterLV-Agent also uses a lightweight two-level memory, where short-term memory stores recent interaction rounds and long-term memory summarizes past observations into compact history notes. Further implementation details are provided in Appendix~\ref{app:agent_details}.

%To standardize evaluation on InterLV-Search, we implement \textbf{InterLV-Agent}, a reference agentic search framework for interleaved language--vision search. InterLV-Agent follows a standard reason-act-observe loop, where a model iteratively plans the next action, invokes tools, observes returned evidence, and updates its search state until producing the final answer.

%The framework provides a unified tool interface covering the capabilities required by InterLV-Search, including image search, reverse image search, web search, webpage browsing, image cropping, and code execution. For Level~1 and Level~2, we further provide an offline multimodal retriever, enabling controlled retrieval over the benchmark corpus. To support long-horizon search, InterLV-Agent maintains a lightweight two-level memory: short-term memory keeps the most recent interaction rounds, while long-term memory compresses past observations into concise summaries of history interaction. The framework also includes budget-aware reflection, trajectory logging, and standardized interfaces for fair model comparison. Further implementation details are provided in Appendix~\ref{app:agent_details}.

\section{Experiment}
\subsection{Experimental Settings}
Following prior work~\citep{zhang2026browsecomp,visbrowse}, we evaluate a diverse set of MLLMs on InterLV-Search, including proprietary general-purpose models (GPT-5.4~\citep{openai2026gpt54thinking}, Gemini-3.1-Pro~\citep{googledeepmind2026gemini31pro}, Claude-Sonnet-4.6~\citep{anthropic2026claudesonnet46}, GPT-5~\citep{singh2025openaigpt5systemcard}, and Qwen3.6-Plus~\citep{qwen2026qwen36plus}) and open-source search-oriented agents on  4$\times$ NVIDIA H20 GPUs (SenseNova-Mars-32B~\citep{chng2025sensenova}, Vision-DeepResearch-8B~\citep{vdr}, and MMSearch-R1~\citep{wu2025mmsearch}). All models are evaluated under the same InterLV-Agent protocol. We report final-answer accuracy as the metric and, following~\citep{java2026characterizing},  use GPT-5.4-mini~\citep{openai2026gpt54thinking} to judge semantic equivalence between model outputs and ground-truth answers, allowing aliases, paraphrases, and minor formatting variations. Following VisBrowse~\citep{visbrowse}, we impose level-specific budgets of 3, 7, and 10 interactions for Level~1, Level~2, and Level~3, respectively, where each interaction includes context observation, tool-call generation, and tool-observation feedback. For Level~3, if multiple tool calls are issued in one interaction, we execute at most the first three to allow parallel search while keeping budgets comparable.

\subsection{Main Results}

Table~\ref{tab:main_results} reports the main results on InterLV-Search. We summarize three observations. 1) \textbf{Tool-free performance confirms the necessity of search.}
Without tools, all models achieve limited accuracy, especially on Level~3, where the best model reaches only 20.00\%. This indicates that InterLV-Search cannot be reliably solved from parametric knowledge alone. % Models must actively retrieve and integrate external visual and textual evidence to answer the questions. %Comparing Direct and +Tool results shows that models differ not only in parametric knowledge, but also in their ability to exploit search. 
2) \textbf{Tool-use gains reveal differences in search capability.} With tool use, proprietary models achieve consistent gains using InterLV-Agent, especially at Level~2 and Level~3, confirming that external evidence acquisition is essential for InterLV-Search. However, performance variance across models is substantial, reflecting differences in their ability to chain tool use for interleaved multimodal evidence search. In contrast, tool use does not bring substantial gains for open-source search-oriented agents, and in some cases even leads to performance degradation. This suggests that existing search-oriented agents remain limited in search planning, visual grounding, and multimodal evidence integration. %use retrieved evidence as a search pivot, decide what to search next, and recover from distractors or wrong branches. This explains the large variation across models. 
%Gemini-3.1-Pro achieves the best +Tool performance across all three levels. GPT-5.4 remains consistently competitive, while Claude-Sonnet-4.6 yields the largest gain on Level~3, improving from 10.50\% to 40.65\% (+30.15). 
%These improvements reflect these models’ ability to plan searches, acquire relevant evidence, and integrate multimodal information.
3) \textbf{Cross-level trends support the staged capability design.}
%The results show a consistent capability dependency across levels. 
%This supports our staged design: Level~1 captures the prerequisite ability of active visual seeking, Level~2 tests controlled evidence-to-query composition, and Level~3 evaluates the same mechanism in the open web. 
Overall, performance decreases from Level~1 to Level~3, reflecting increasing difficulty in interleaved multimodal search, which supports our staged benchmark design. Notably, some Level~3 +Tool scores exceed those on Level~2, since Level~3 provides a larger interaction budget and allows more recovery attempts, rather than being intrinsically easier. Within Level~3, all models perform substantially worse on multi-branch examples than on single-chain ones. This indicates that current agents are less robust to complex search topologies, highlighting the value of InterLV-Search in evaluating non-linear multimodal search control. % beyond single-chain multi-hop reasoning.

\newcommand{\pos}[1]{\textcolor{green!60!black}{+#1}}
\newcommand{\negg}[1]{\textcolor{red}{#1}}

% Required packages:
% \usepackage{booktabs}
% \usepackage{multirow}
% \usepackage{makecell}
% \usepackage{xcolor}

\begin{table*}[t]
\centering
\small
\setlength{\tabcolsep}{3.0pt}
\renewcommand{\arraystretch}{1.08}

\begingroup
\setlength{\heavyrulewidth}{0.9pt}
\setlength{\lightrulewidth}{0.45pt}
\setlength{\cmidrulewidth}{0.45pt}
\setlength{\arrayrulewidth}{0.30pt}

\caption{
Main results on InterLV-Search in terms of final-answer accuracy (\%).
Direct denotes direct answering; +Tool denotes InterLV-Agent evaluation; $\Delta$ is the accuracy change from tool use. 
%For Level~3, Single and Multi denote accuracy on single-chain and multi-branch examples, respectively.
}
\label{tab:main_results}

\begin{tabular}{lccc|ccc|ccccc}
\toprule
\multirow{2}{*}{Model}
& \multicolumn{3}{c|}{Level~1}
& \multicolumn{3}{c|}{Level~2}
& \multicolumn{5}{c}{Level~3} \\
\cmidrule(lr){2-4} \cmidrule(lr){5-7} \cmidrule(lr){8-12}
& Direct & \makecell{+Tool} & $\Delta$
& Direct & \makecell{+Tool} & $\Delta$
& Direct & \makecell{+Tool\\Avg.} & \makecell{+Tool\\Single}& \makecell{+Tool \\Multi} & $\Delta$ \\
\midrule
\multicolumn{12}{l}{\textit{Proprietary General MLLMs}} \\

GPT-5
& 32.14 & 39.54 & \pos{7.40}
& 20.00 & 35.11 & \pos{15.11}
& 8.60 & 35.19 & 40.12 & 27.65 & \pos{26.59} \\

GPT-5.4
& 35.96 & 45.64 & \pos{9.68}
& 27.44 & 38.22 & \pos{10.78}
& \textbf{20.00} & 44.25 & 51.06 & 33.82 & \pos{24.25} \\

Claude-Sonnet-4.6
& 28.51 & 37.13 & \pos{8.62}
& \textbf{30.89} & 34.22 & \pos{3.33}
& 10.50 & 40.65 & 46.83 & 33.18 & \pos{30.15} \\

Gemini-3.1-Pro
& \textbf{41.29} & \textbf{46.05} & \pos{4.76}
& 28.52 & \textbf{41.33} & \pos{12.81}
& 17.20 & \textbf{46.46} & \textbf{52.02} & \textbf{37.94}  & \pos{29.26} \\

Qwen3.6-Plus
& 22.44 & 29.27 & \pos{6.83}
& 22.44 & 27.56 & \pos{5.11}
& 10.76 & 37.51 & 42.80 & 29.41 & \pos{26.75} \\

\midrule
\multicolumn{12}{l}{\textit{Open-Source Search-oriented MLLMs}} \\

MMSearch-R1-7B
& 5.94 & 4.10 & \negg{-1.85}
& 7.11 & 5.33 & \negg{-1.78}
& 4.77 & 11.96 & 7.60 & 14.77 & \pos{7.19} \\

VDR-8B
& 2.77 & 3.13 & \pos{0.36}
& 8.00 & 6.90 & \negg{-1.10}
& 5.46 & 15.56 & 18.04 & 11.76 & \pos{10.10} \\

SenseNova-MARS-32B
& 19.28 & 15.69 & \negg{-3.59}
& 15.56 & 10.67 & \negg{-4.89}
& 6.56 & 29.73 & 34.93 & 21.76 & \pos{23.17} \\

\bottomrule
\end{tabular}
%\vspace{-1.5em}
\endgroup

\end{table*}

\subsection{Target Retrieval--Answer Decomposition for Levels 1 and 2}
%InterLV-Search allows us to decompose performance on target entity retrieval and final question answering at
According to the construction paradigms of Level~1 and Level~2, we know the final target entities. %This enables evaluation not only of final answer correctness, but also of whether the intended visual evidence is retrieved.
We thus conduct the decomposition analysis over the final target entity retrieval and final answer correctness with the three strongest models in our main results: Gemini-3.1-Pro, GPT-5.4, and Claude-Sonnet-4.6. We report retrieval recall (Ret.~R@5) by checking whether the top-5 entities returned in the final retrieval step contain the ground-truth target entity; and report answer accuracy under three settings: overall accuracy (Acc.), accuracy when the target entity is successfully retrieved (Acc.$\mid$Ret.), and accuracy when it is not retrieved (Acc.$\mid$UnRet.). We also report Corr.~from Ret., which measures the fraction of correct answers accompanied by successful target retrieval.

%We report retrieval recall, Ret.~R@5, by checking whether the top-5 entities returned in the model's final retrieval step contain the ground-truth target entity; answer accuracy under three settings: overall accuracy (Acc.), accuracy when the target entity is successfully retrieved (Acc.$\mid$Ret.), and accuracy when it is not retrieved (Acc.$\mid$UnRet.), as well as Corr.~from Ret., which measures the fraction of correct answers that are accompanied by successful target retrieval. 
%The analysis further distinguishes evidence-grounded answers from those obtained via guessing, shortcuts, or incidental cues.
%InterLV-Search allows us to decompose performance on Level~1 and Level~2 because both levels are constructed with known target entities and terminal VQA questions. We can therefore check not only whether the final answer is correct, but also whether the agent actually retrieves the intended visual evidence. This analysis helps distinguish answers supported by correct evidence localization from answers obtained through guessing, shortcuts, or incidental information.

%This metric is important because high Acc.$\mid$Ret. alone can be misleading: a model may answer retrieved cases accurately, but if it retrieves the target in only a small number of examples, most of its correct answers may still come from unretrieved cases. Such answers are less likely to be supported by the intended visual evidence and may instead reflect guessing, shortcuts, or incidental evidence.

As shown in Table~\ref{tab:retrieval_decomposition}, Acc.$\mid$Ret. is consistently higher than Acc.$\mid$UnRet., especially on Level~2, showing that agents can often answer correctly once the intended visual evidence is retrieved. Corr.~from Ret. remains high even with relatively low Ret.~R@5, further indicating that successful retrieval contributes a substantial share of correct answers. However, the limited Ret.~R@5, particularly on Level~2, indicates that target evidence localization is still a major bottleneck.

\begin{table*}[t]
\centering
\small
\setlength{\tabcolsep}{3.6pt}
\renewcommand{\arraystretch}{1.10}
\caption{
Retrieval--answer decomposition results on Level~1 and Level~2. 
All values are percentages.
%Ret. R@5 denotes top-5 recall of the target visual evidence. Ret. Acc. and UnRet. Acc. denote answer accuracy conditioned on whether the target evidence is retrieved. Corr. from Ret. denotes the percentage of correct answers whose target evidence is retrieved. 
}
\label{tab:retrieval_decomposition}
\begin{tabular}{lccccc|ccccc}
\toprule
\multirow{2}{*}{Model}
& \multicolumn{5}{c|}{Level~1: Active Visual Evidence Seeking}
& \multicolumn{5}{c}{Level~2: Offline Interleaved Search} \\
\cmidrule(lr){2-6} \cmidrule(lr){7-11}
& \makecell{Ret.\\R@5}
& Acc.
& \makecell{Ret.\\Acc.}
& \makecell{UnRet.\\Acc.}
& \makecell{Corr.\\from Ret.}
& \makecell{Ret.\\R@5}
& Acc.
& \makecell{Ret.\\Acc.}
& \makecell{UnRet.\\Acc.}
& \makecell{Corr.\\from Ret.} \\
\midrule
Gemini-3.1-Pro
& 46.36 & \textbf{46.05} & \textbf{59.51} & \textbf{34.42} & 59.91
& \textbf{35.56} & \textbf{41.33} & \textbf{73.75} & 23.45 & \textbf{63.44} \\

GPT-5.4
& \textbf{53.95} & 45.64 & 58.17 & 30.96 & \textbf{68.76}
& 31.56 & 38.22 & 64.79 & \textbf{25.97} & 53.49 \\

Claude-Sonnet-4.6
& 35.59 & 37.13 & 56.77 & 26.27 & 54.42
& 21.33 & 34.22 & 72.92 & 23.73 & 45.45 \\
\bottomrule
\end{tabular}

\end{table*}

\subsection{Further Analysis}
\begin{table*}[t]
\centering
\small
\renewcommand{\arraystretch}{1.08}

\begingroup
\setlength{\heavyrulewidth}{0.9pt}
\setlength{\lightrulewidth}{0.45pt}
\setlength{\cmidrulewidth}{0.45pt}
\setlength{\arrayrulewidth}{0.30pt}
\caption{
Ablation on interleaved-search components for Level~2 and Level~3.
}
\label{tab:level_ablation}

\resizebox{\linewidth}{!}{%

\begin{tabular}{lccc|ccc}
\toprule
\multirow{2}{*}{Setting}
& \multicolumn{3}{c|}{Level~2: Offline Interleaved Search}
& \multicolumn{3}{c}{Level~3: Open-Web Interleaved Search} \\
\cmidrule(lr){2-4} \cmidrule(lr){5-7}
& GPT-5.4 & Gemini-3.1-Pro & Claude-4.6-Sonnet
& GPT-5.4 & Gemini-3.1-Pro & Claude-4.6-Sonnet \\
\midrule
Direct 
& 27.44 & 28.52 & 30.89
& 20.00 & 17.20 & 10.50 \\

w/o Image Search 
& 28.89 & 27.22 & 24.00
& 36.12 & 38.91 & 35.20 \\

w/o Memory 
& 36.89 & 40.00 & \textbf{35.55}
& 40.42 & 44.48 & 37.63 \\

Full 
& \textbf{38.22} & \textbf{41.33} & 34.22
& \textbf{44.25} & \textbf{46.46} & \textbf{40.65} \\
\bottomrule
\end{tabular}%
}

\endgroup

\vspace{-1em}
\end{table*}%\textbf{What are the key capabilities of Open-Web Interleave Search?}   Table~\ref{tab:level3_ablation} shows Level~3 ablations. Removing image search consistently reduces accuracy, confirming that Level~3 requires active visual retrieval rather than text-only browsing. Disabling running memory also hurts all three models. Since Level~3 is designed around interleaved trajectories where visual evidence serves as a search pivot, agents must retain intermediate entities, visual clues, branch states, and unresolved subgoals across multiple search steps. Without memory, these signals are easily diluted by noisy open-web observations, making it harder to decide what to search next. Together, these results support the design of InterLV-Search: solving the benchmark requires both visual evidence seeking and long-horizon interleaved search, rather than isolated visual lookup or shallow browsing.
\begin{figure*}[h]
    \centering
    \includegraphics[width=\textwidth]{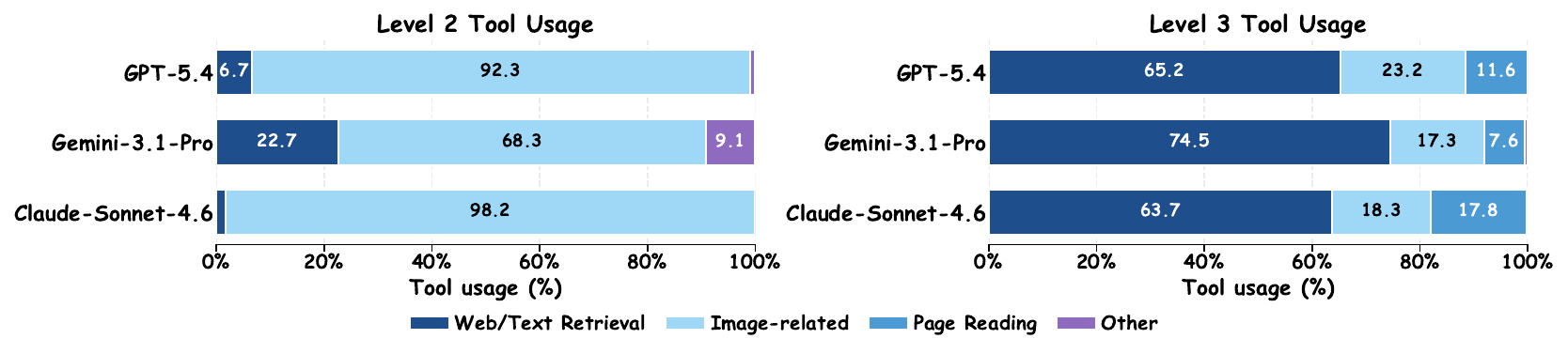}
    \vspace{-1em}

    \caption{
    Tool-usage distribution on Level~2 and Level~3. 
   % Level~2 is dominated by image-related retrieval, while Level~3 combines web/text retrieval, page reading, and image-related retrieval.
}
    
    \label{fig:tool_usage}
    \vspace{-1em}
\end{figure*}
\textbf{What capabilities are required by interleaved search?}
Table~\ref{tab:level_ablation} reports component ablations on Level~2 and Level~3. On Level~2, removing image search leads to a significant performance drop because the offline environment is constrained and evidence is sparse; image search is often essential for locating key visual pivots. Without it, the model fails to retrieve critical evidence and can even underperform the Direct (no-tool) baseline due to ineffective search paths. On Level~3, the impact is smaller, likely because the web setting provides richer textual evidence that can partially substitute visual signals. However, performance still consistently degrades, indicating that visual retrieval remains beneficial even in noisy web environments. Memory shows a clearer effect on Level~3 than Level~2, which is likely because Level~2 chains are relatively shorter, while Level~3 typically involves longer trajectories, branch exploration, and noisier observations. As a result, agents need to maintain memory of intermediate entities, visual cues, and unresolved subgoals to decide what to search next.% Overall, the ablation suggests that InterLV-Search requires both precise visual evidence retrieval and memory maintenance for long-horizon interleaved search. % rather than text-only browsing or isolated image lookup.

\textbf{What tools do agents actually use?}
Fig.~\ref{fig:tool_usage} shows the tool usage distribution of top-performing models. As shown, Level~2 is dominated by image-related retrieval, consistent with its visual–entity transition design: agents must retrieve visual evidence and use it to guide subsequent search. Level~3 relies more on web/text retrieval, as the open web provides diverse but noisy sources for evidence search. Nevertheless, image-related tools\footnote{
Image-related tools include all image-centric retrieval or inspection operations, including local image retrieval, local text-to-image retrieval, online image search, reverse image search, screenshot browsing, and image cropping.
}
still account for a substantial fraction of calls, showing that Level~3 does not reduce to text-only web browsing. %We also observe model-specific preferences: GPT-5.4 and Claude-Sonnet-4.6 use image-related tools more frequently, explaining their larger tool-use gains, while Gemini-3.1-Pro relies more on web/text retrieval, likely because stronger parametric and web-search reasoning helps narrow candidate sources before invoking visual tools.

\begin{wrapfigure}{r}{0.4\textwidth}
    
    \centering
    \vspace{-9pt}
    \includegraphics[width=0.4\textwidth]{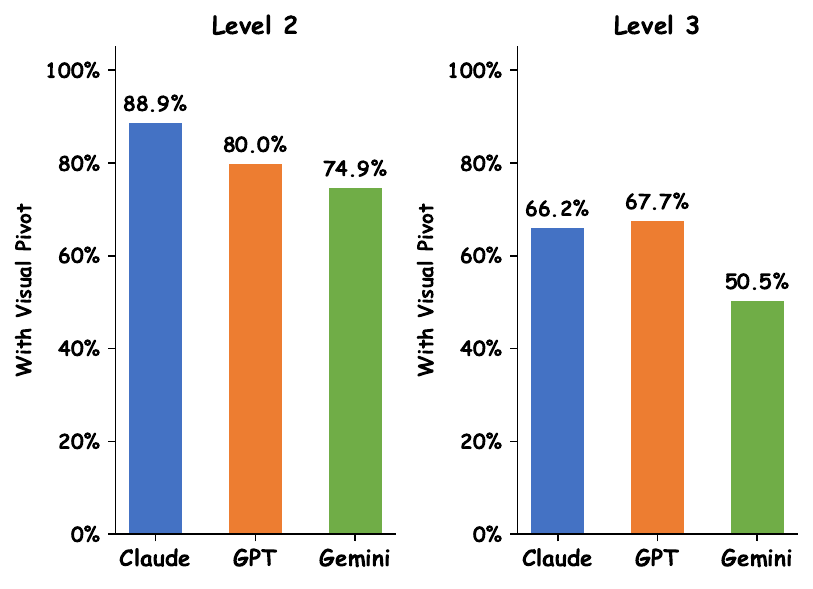}
    \vspace{-0.8em}
    \caption{
Fraction of examples with visual pivots in executed trajectories.    }
    \vspace{-1em}
    \label{fig:visual_chain_ratio}
    \vspace{-5pt}
    %\vspace{-1em}
\end{wrapfigure}
\textbf{Do model trajectories actually contain visual pivots?}
To directly verify whether agents follow the intended interleaved pattern, we analyze the logged trajectories with an LLM-based trajectory judge. 
Following our definition of search-controlling visual evidence, we define a \emph{visual-pivot trajectory} as one where visual evidence is used to guide subsequent search rather than only final answering.  As shown in Fig.~\ref{fig:visual_chain_ratio}, a large fraction of executed trajectories contain visual pivots. 
The ratio is especially high on Level~2 and remains substantial on Level~3 despite open-web noise. 
This provides trajectory-level evidence that InterLV-Search does require agents to use visual evidence inside the search process to guide subsequent retrieval.

%We further analyse Tool Usage of three models. As shown in Fig.~\ref{fig:tool_usage}, Level~2 is dominated by image-related retrieval, reflecting its construction around visual-entity transitions. Agents must locate visual evidence and use the retrieved visual entity as a search pivot for subsequent steps, confirming that Level~2 evaluates controlled interleaved search rather than text-only multi-hop retrieval.

%Figure~\ref{fig:visual_chain_ratio} further shows that the executed trajectories frequently contain visual chains. On Level~2, 74.9--88.9\% of trajectories include a visual chain across the three top-performing models, consistent with its controlled construction around visual-entity transitions. On Level~3, the ratio remains substantial despite the noisier open-web environment, reaching 50.5--67.7\%. This confirms that InterLV-Search does not merely add isolated image lookup; visual evidence is repeatedly used inside the search trajectory, supporting our goal of evaluating interleaved multimodal search.
%\vspace{-1em}
\begin{figure*}[h]
    \centering
    \includegraphics[width=\textwidth]{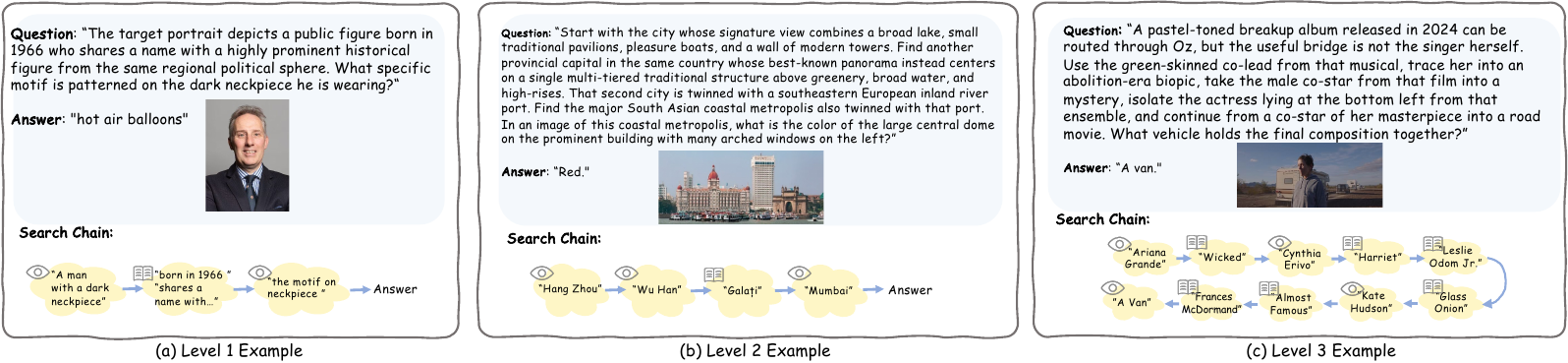}
    \caption{InterLV-Search question-answer examples with plausible evidence paths.}
%, each paired with a reference search chain illustrating the plausible evidence path.}
%each consisting of a question, answer, and search chain that illustrates the intended evidence path and required search logic. }
    \label{fig:case_study}
    \vspace{-1em}
\end{figure*}

\subsection{Case Study}
Figure~\ref{fig:case_study} shows representative question–answer examples along with expected search chains across the three levels of InterLV-Search. Level~1 starts from a textual description of a visual cue and requires active retrieval of the target image before answering. Level~2 follows an interleaved chain, such as Hangzhou (Visual) $\rightarrow$ Wuhan (Visual)  $\rightarrow$ Galati  (Textual) $\rightarrow$ Mumbai (Visual), where retrieved visual evidence acts as pivots for subsequent entity transitions. 
Level~3 extends this pattern to the open web, where the chain may traverse album, film, actor, and movie evidence across noisier sources. More detailed discussions are provided in Appendix~\ref{app:case_study}.

%These cases show that InterLV-Search evaluates not only whether agents can find visual evidence, but whether they can use it to guide later search decisions. 
%In Level~1, the agent starts from a textual cue and must actively seek the relevant image before answering, testing the primitive ability of visual evidence acquisition. 

%In Level~2, the search becomes genuinely interleaved: the agent first follows a textual chain, then uses retrieved visual evidence as a \emph{search pivot} to determine the next entity or retrieval target, before finally answering a terminal visual question. In Level~3, the same interleaved mechanism is required in the open web, where the agent must navigate longer and noisier trajectories and integrate evidence from multiple sources. 

\section{Conclusion}
This work shows that interleaved multimodal search exposes challenges not captured by existing benchmarks for agentic search. %endpoint-oriented visual browsing or text-centric search evaluation. 
Across benchmark levels, retrieval--answer decomposition, tool-use analysis, and trajectory inspection consistently validate the design of InterLV-Search: success depends not only on accessing external tools, but on locating the intended visual evidence, using it as a search pivot, and maintaining coherent search state across long or branching trajectories. The substantial performance gaps across levels, between single-chain and multi-branch tasks, and between retrieved and unretrieved cases demonstrate that current multimodal agents remain far from robust open-world interleaved search. We hope InterLV-Search can support future work on agents that more reliably acquire, connect, and act on multimodal evidence during search.
\clearpage
\newpage
\bibliographystyle{assets/plainnat}
\bibliography{paper}

@article{comanici2025gemini,
  title={Gemini 2.5: Pushing the frontier with advanced reasoning, multimodality, long context, and next generation agentic capabilities},
  author={Comanici, Gheorghe and Bieber, Eric and Schaekermann, Mike and Pasupat, Ice and Sachdeva, Noveen and Dhillon, Inderjit and Blistein, Marcel and Ram, Ori and Zhang, Dan and Rosen, Evan and others},
  journal={arXiv preprint arXiv:2507.06261},
  year={2025}
}

@inproceedings{li2025search,
  title={Search-o1: Agentic search-enhanced large reasoning models},
  author={Li, Xiaoxi and Dong, Guanting and Jin, Jiajie and Zhang, Yuyao and Zhou, Yujia and Zhu, Yutao and Zhang, Peitian and Dou, Zhicheng},
  booktitle={Proceedings of the 2025 Conference on Empirical Methods in Natural Language Processing},
  pages={5420--5438},
  year={2025}
}

@article{bai2025qwen3,
  title={Qwen3-vl technical report},
  author={Bai, Shuai and Cai, Yuxuan and Chen, Ruizhe and Chen, Keqin and Chen, Xionghui and Cheng, Zesen and Deng, Lianghao and Ding, Wei and Gao, Chang and Ge, Chunjiang and others},
  journal={arXiv preprint arXiv:2511.21631},
  year={2025}
}

@article{team2026kimi,
  title={Kimi K2. 5: Visual Agentic Intelligence},
  author={Team, Kimi and Bai, Tongtong and Bai, Yifan and Bao, Yiping and Cai, SH and Cao, Yuan and Charles, Y and Che, HS and Chen, Cheng and Chen, Guanduo and others},
  journal={arXiv preprint arXiv:2602.02276},
  year={2026}
}

@article{jiang2024mmsearch,
  title={Mmsearch: Benchmarking the potential of large models as multi-modal search engines},
  author={Jiang, Dongzhi and Zhang, Renrui and Guo, Ziyu and Wu, Yanmin and Lei, Jiayi and Qiu, Pengshuo and Lu, Pan and Chen, Zehui and Fu, Chaoyou and Song, Guanglu and others},
  journal={arXiv preprint arXiv:2409.12959},
  year={2024}
}

@article{guo2025deepseek,
  title={Deepseek-r1: Incentivizing reasoning capability in llms via reinforcement learning},
  author={Guo, Daya and Yang, Dejian and Zhang, Haowei and Song, Junxiao and Wang, Peiyi and Zhu, Qihao and Xu, Runxin and Zhang, Ruoyu and Ma, Shirong and Bi, Xiao and others},
  journal={arXiv preprint arXiv:2501.12948},
  year={2025}
}

@article{wu2025mmsearch,
  title={Mmsearch-r1: Incentivizing lmms to search},
  author={Wu, Jinming and Deng, Zihao and Li, Wei and Liu, Yiding and You, Bo and Li, Bo and Ma, Zejun and Liu, Ziwei},
  journal={arXiv preprint arXiv:2506.20670},
  year={2025}
}

@inproceedings{
geng2026webwatcher,
title={WebWatcher: Breaking New Frontiers of Vision-Language Deep Research Agent},
author={Xinyu Geng and Peng Xia and Zhen Zhang and Xinyu Wang and Qiuchen Wang and Ruixue Ding and Chenxi Wang and Jialong Wu and Kuan Li and Yida Zhao and Huifeng Yin and Yong Jiang and Pengjun Xie and Fei Huang and Huaxiu Yao and Yi R. Fung and Jingren Zhou},
booktitle={The Fourteenth International Conference on Learning Representations},
year={2026},
url={https://openreview.net/forum?id=8jsaazdAb3}
}

@article{zeng2026vision,
  title={Vision-DeepResearch Benchmark: Rethinking Visual and Textual Search for Multimodal Large Language Models},
  author={Zeng, Yu and Huang, Wenxuan and Fang, Zhen and Chen, Shuang and Shen, Yufan and Cai, Yishuo and Wang, Xiaoman and Yin, Zhenfei and Chen, Lin and Chen, Zehui and others},
  journal={arXiv preprint arXiv:2602.02185},
  year={2026}
}

@misc{visbrowse,
   title={VisBrowse-Bench: Benchmarking Visual-Native Search for Multimodal Browsing Agents}, 
   author={Zhengbo Zhang and Jinbo Su and Zhaowen Zhou and Changtao Miao and Yuhan Hong and Qimeng Wu and Yumeng Liu and Feier Wu and Yihe Tian and Yuhao Liang and Zitong Shan and Wanke Xia and Yi-Fan Zhang and Bo Zhang and Zhe Li and Shiming Xiang and Ying Yan},
   year={2026},
   eprint={2603.16289},
   archivePrefix={arXiv},
   primaryClass={cs.CV},
   url={https://arxiv.org/abs/2603.16289}, 
}

@article{narayan2025deepmmsearch,
  title={Deepmmsearch-r1: Empowering multimodal llms in multimodal web search},
  author={Narayan, Kartik and Xu, Yang and Cao, Tian and Nerella, Kavya and Patel, Vishal M and Shiee, Navid and Grasch, Peter and Jia, Chao and Yang, Yinfei and Gan, Zhe},
  journal={arXiv preprint arXiv:2510.12801},
  year={2025}
}

@inproceedings{
zhou2024webarena,
title={WebArena: A Realistic Web Environment for Building Autonomous Agents},
author={Shuyan Zhou and Frank F. Xu and Hao Zhu and Xuhui Zhou and Robert Lo and Abishek Sridhar and Xianyi Cheng and Tianyue Ou and Yonatan Bisk and Daniel Fried and Uri Alon and Graham Neubig},
booktitle={The Twelfth International Conference on Learning Representations},
year={2024},
url={https://openreview.net/forum?id=oKn9c6ytLx}
}

@inproceedings{
du2026deepresearch,
title={DeepResearch Bench: A Comprehensive Benchmark for Deep Research Agents},
author={Mingxuan Du and Benfeng Xu and Chiwei Zhu and Licheng Zhang and Xiaorui Wang and Zhendong Mao},
booktitle={The Fourteenth International Conference on Learning Representations},
year={2026},
url={https://openreview.net/forum?id=hQ0K2Hhq7H}
}

@inproceedings{roberts2020much,
  title={How much knowledge can you pack into the parameters of a language model?},
  author={Roberts, Adam and Raffel, Colin and Shazeer, Noam},
  booktitle={Proceedings of the 2020 conference on empirical methods in natural language processing (EMNLP)},
  pages={5418--5426},
  year={2020}
}

@inproceedings{goyal2017making,
  title={Making the v in vqa matter: Elevating the role of image understanding in visual question answering},
  author={Goyal, Yash and Khot, Tejas and Summers-Stay, Douglas and Batra, Dhruv and Parikh, Devi},
  booktitle={Proceedings of the IEEE conference on computer vision and pattern recognition},
  pages={6904--6913},
  year={2017}
}

@inproceedings{ye2023enhancing,
  title={Enhancing conversational search: Large language model-aided informative query rewriting},
  author={Ye, Fanghua and Fang, Meng and Li, Shenghui and Yilmaz, Emine},
  booktitle={Findings of the Association for Computational Linguistics: EMNLP 2023},
  pages={5985--6006},
  year={2023}
}

@inproceedings{faggioli2024query,
  title={Query obfuscation for information retrieval through differential privacy},
  author={Faggioli, Guglielmo and Ferro, Nicola},
  booktitle={European Conference on Information Retrieval},
  pages={278--294},
  year={2024},
  organization={Springer}
}

@article{huang2026mmdeepresearch,
  title={MMDeepResearch-Bench: A Benchmark for Multimodal Deep Research Agents},
  author={Huang, Peizhou and Zhong, Zixuan and Wan, Zhongwei and Zhou, Donghao and Alam, Samiul and Wang, Xin and Li, Zexin and Dou, Zhihao and Zhu, Li and Xiong, Jing and others},
  journal={arXiv preprint arXiv:2601.12346},
  year={2026}
}

@article{li2025mm,
  title={MM-browsecomp: A comprehensive benchmark for multimodal browsing agents},
  author={Li, Shilong and Bu, Xingyuan and Wang, Wenjie and Liu, Jiaheng and Dong, Jun and He, Haoyang and Lu, Hao and Zhang, Haozhe and Jing, Chenchen and Li, Zhen and others},
  journal={arXiv preprint arXiv:2508.13186},
  year={2025}
}

@article{wei2025browsecomp,
  title={Browsecomp: A simple yet challenging benchmark for browsing agents},
  author={Wei, Jason and Sun, Zhiqing and Papay, Spencer and McKinney, Scott and Han, Jeffrey and Fulford, Isa and Chung, Hyung Won and Passos, Alex Tachard and Fedus, William and Glaese, Amelia},
  journal={arXiv preprint arXiv:2504.12516},
  year={2025}
}

@article{zhang2026browsecomp,
  title={BrowseComp-$ V3$: A Visual, Vertical, and Verifiable Benchmark for Multimodal Browsing Agents},
  author={Zhang, Huanyao and Zhou, Jiepeng and Li, Bo and Zhou, Bowen and Shan, Yanzhe and Lu, Haishan and Cao, Zhiyong and Chen, Jiaoyang and Han, Yuqian and Sheng, Zinan and others},
  journal={arXiv preprint arXiv:2602.12876},
  year={2026}
}

@inproceedings{luo2021weakly,
  title={Weakly-supervised visual-retriever-reader for knowledge-based question answering},
  author={Luo, Man and Zeng, Yankai and Banerjee, Pratyay and Baral, Chitta},
  booktitle={Proceedings of the 2021 Conference on Empirical Methods in Natural Language Processing},
  pages={6417--6431},
  year={2021}
}

@inproceedings{
hong2026knowledgebased,
title={Knowledge-based Visual Question Answer with Multimodal Processing, Retrieval and Filtering},
author={Yuyang Hong and Jiaqi Gu and Qi Yang and Lubin Fan and Yue Wu and Ying Wang and Kun Ding and Shiming Xiang and Jieping Ye},
booktitle={The Thirty-ninth Annual Conference on Neural Information Processing Systems},
year={2026},
url={https://openreview.net/forum?id=h0LzGQq6uO}
}

@inproceedings{hou2025fire,
  title={FiRE: Enhancing MLLMs with fine-grained context learning for complex image retrieval},
  author={Hou, Bohan and Lin, Haoqiang and Song, Xuemeng and Wen, Haokun and Liu, Meng and Hu, Yupeng and Zhao, Xiangyu},
  booktitle={Proceedings of the 48th International ACM SIGIR Conference on Research and Development in Information Retrieval},
  pages={803--812},
  year={2025}
}

@article{song2025comprehensive,
  title={A comprehensive survey on composed image retrieval},
  author={Song, Xuemeng and Lin, Haoqiang and Wen, Haokun and Hou, Bohan and Xu, Mingzhu and Nie, Liqiang},
  journal={ACM Transactions on Information Systems},
  volume={44},
  number={1},
  pages={1--54},
  year={2025},
  publisher={ACM New York, NY}
}

@inproceedings{du2025easy,
  title={From Easy to Hard: The MIR Benchmark for Progressive Interleaved Multi-Image Reasoning},
  author={Du, Hang and Zhang, Jiayang and Nan, Guoshun and Deng, Wendi and Chen, Zhenyan and Zhang, Chenyang and Xiao, Wang and Huang, Shan and Pan, Yuqi and Qi, Tao and others},
  booktitle={Proceedings of the IEEE/CVF International Conference on Computer Vision},
  pages={859--869},
  year={2025}
}

@misc{deng2026deepimagesearchbenchmarkingmultimodalagents,
  title={DeepImageSearch: Benchmarking Multimodal Agents for Context-Aware Image Retrieval in Visual Histories}, 
  author={Chenlong Deng and Mengjie Deng and Junjie Wu and Dun Zeng and Teng Wang and Qingsong Xie and Jiadeng Huang and Shengjie Ma and Changwang Zhang and Zhaoxiang Wang and Jun Wang and Yutao Zhu and Zhicheng Dou},
  year={2026},
  eprint={2602.10809},
  archivePrefix={arXiv},
  primaryClass={cs.CV},
  url={https://arxiv.org/abs/2602.10809}
}

@inproceedings{perron2023cpsatlp,
  author    = {Perron, Laurent and Didier, Fr{\'e}d{\'e}ric and Gay, Steven},
  title     = {The {CP-SAT-LP} Solver},
  booktitle = {29th International Conference on Principles and Practice of Constraint Programming (CP 2023)},
  pages     = {3:1--3:2},
  series    = {Leibniz International Proceedings in Informatics (LIPIcs)},
  volume    = {280},
  year      = {2023},
  publisher = {Schloss Dagstuhl -- Leibniz-Zentrum f{\"u}r Informatik},
  doi       = {10.4230/LIPIcs.CP.2023.3},
  url       = {https://drops.dagstuhl.de/entities/document/10.4230/LIPIcs.CP.2023.3}
}

@inproceedings{
zhang2025mmkg,
title={Multiple Heads are Better than One: Mixture of Modality Knowledge Experts for Entity Representation Learning},
author={Yichi Zhang and Zhuo Chen and Lingbing Guo and yajing Xu and Binbin Hu and Ziqi Liu and Wen Zhang and Huajun Chen},
booktitle={The Thirteenth International Conference on Learning Representations},
year={2025},
url={https://openreview.net/forum?id=ue1Tt3h1VC}
}

@article{chng2025sensenova,
  title={SenseNova-MARS: Empowering Multimodal Agentic Reasoning and Search via Reinforcement Learning},
  author={Chng, Yong Xien and Hu, Tao and Tong, Wenwen and Li, Xueheng and Chen, Jiandong and Yu, Haojia and Lu, Jiefan and Guo, Hewei and Deng, Hanming and Xie, Chengjun and others},
  journal={arXiv preprint arXiv:2512.24330},
  year={2025}
}

@article{vdr,
  title={Vision-deepresearch: Incentivizing deepresearch capability in multimodal large language models},
  author={Huang, Wenxuan and Zeng, Yu and Wang, Qiuchen and Fang, Zhen and Cao, Shaosheng and Chu, Zheng and Yin, Qingyu and Chen, Shuang and Yin, Zhenfei and Chen, Lin and others},
  journal={arXiv preprint arXiv:2601.22060},
  year={2026}
}

@inproceedings{chang2022webqa,
  title={Webqa: Multihop and multimodal qa},
  author={Chang, Yingshan and Narang, Mridu and Suzuki, Hisami and Cao, Guihong and Gao, Jianfeng and Bisk, Yonatan},
  booktitle={Proceedings of the IEEE/CVF conference on computer vision and pattern recognition},
  pages={16495--16504},
  year={2022}
}

@inproceedings{yao2023react,
  title = {{ReAct}: Synergizing Reasoning and Acting in Language Models},
  author = {Yao, Shunyu and Zhao, Jeffrey and Yu, Dian and Du, Nan and Shafran, Izhak and Narasimhan, Karthik and Cao, Yuan},
  booktitle = {International Conference on Learning Representations (ICLR) },
  year = {2023},
  html = {https://arxiv.org/abs/2210.03629},
}

@inproceedings{koh2024visualwebarena,
  title={Visualwebarena: Evaluating multimodal agents on realistic visual web tasks},
  author={Koh, Jing Yu and Lo, Robert and Jang, Lawrence and Duvvur, Vikram and Lim, Ming and Huang, Po-Yu and Neubig, Graham and Zhou, Shuyan and Salakhutdinov, Russ and Fried, Daniel},
  booktitle={Proceedings of the 62nd Annual Meeting of the Association for Computational Linguistics (Volume 1: Long Papers)},
  pages={881--905},
  year={2024}
}

@inproceedings{
tao2026mmsearchplus,
title={{MMS}earch-Plus: Benchmarking Provenance-Aware Search for Multimodal Browsing Agents},
author={Xijia Tao and Teng Yihua and Xinxing Su and Xinyu Fu and Jihao Wu and Chaofan Tao and Ziru Liu and Haoli Bai and Rui Liu and Lingpeng Kong},
booktitle={The Fourteenth International Conference on Learning Representations},
year={2026},
url={https://openreview.net/forum?id=VGYgG2GH0d}
}

@inproceedings{marino2019ok,
  title={Ok-vqa: A visual question answering benchmark requiring external knowledge},
  author={Marino, Kenneth and Rastegari, Mohammad and Farhadi, Ali and Mottaghi, Roozbeh},
  booktitle={Proceedings of the IEEE/cvf conference on computer vision and pattern recognition},
  pages={3195--3204},
  year={2019}
}

@inproceedings{
java2026characterizing,
title={Characterizing Deep Research: A Benchmark and Formal Definition},
author={Abhinav Java and Ashmit Khandelwal and Sukruta Prakash Midigeshi and Aaron Halfaker and Amit Deshpande and Navin Goyal and Ankur Gupta and Nagarajan Natarajan and Amit Sharma},
booktitle={The Fourteenth International Conference on Learning Representations},
year={2026},
url={https://openreview.net/forum?id=5EmpOCq1Ql}
}

@misc{openai2026gpt54thinking,
  title        = {{GPT-5.4 Thinking System Card}},
  author       = {{OpenAI}},
  year         = {2026},
  month        = mar,
  howpublished = {https://deploymentsafety.openai.com/gpt-5-4-thinking},
  note         = {Published March 5, 2026}
}

@misc{googledeepmind2026gemini31pro,
  title        = {{Gemini 3.1 Pro Model Card}},
  author       = {{Google DeepMind}},
  year         = {2026},
  month        = feb,
  howpublished = {https://deepmind.google/models/model-cards/gemini-3-1-pro/}
}

@misc{anthropic2026claudesonnet46,
  title        = {{Claude Sonnet 4.6 System Card}},
  author       = {{Anthropic}},
  year         = {2026},
  month        = feb,
  howpublished = {https://www.anthropic.com/claude-sonnet-4-6-system-card}
}

@article{singh2025openaigpt5systemcard,
  title   = {{OpenAI GPT-5 System Card}},
  author  = {Singh, Aaditya and Fry, Adam and Perelman, Adam and Tart, Adam and Ganesh, Adi and El-Kishky, Ahmed and McLaughlin, Aidan and others},
  journal = {arXiv preprint arXiv:2601.03267},
  year    = {2025},
  doi     = {10.48550/arXiv.2601.03267}
}

@misc{qwen2026qwen36plus,
  title        = {{Qwen3.6-Plus}: Towards Real World Agents},
  author       = {{Qwen Team}},
  year         = {2026},
  month        = apr,
  howpublished = {https://qwen.ai/blog?id=qwen3.6}
}
\appendix
\clearpage
\newpage
\setcounter{page}{1}

\startcontents[chapters]

\section*{Appendix}
%\section{More Results}
\printcontents[chapters]{}{1}{}
\section{Related Work}
\subsection{Multimodal Search Agent}

Recent LLMs and MLLMs have greatly improved general reasoning and multimodal understanding~\citep{singh2025openaigpt5systemcard,comanici2025gemini,bai2025qwen3,team2026kimi}. 
However, open-domain tasks often require evidence that is not contained in the input or model parameters. 
This has motivated multimodal search agents, which extend static multimodal reasoning with tool use, web browsing, image search, and iterative evidence acquisition~\citep{jiang2024mmsearch,wu2025mmsearch,li2025search}. 
Instead of answering directly from a fixed context, these agents decompose the user query into subgoals, issue search or browsing actions, inspect returned textual and visual evidence, and update their next actions based on new observations.

Existing multimodal search agents have shown promising ability in web-assisted question answering and tool-augmented reasoning, but their search behavior is still often text-centric. 
Visual information is commonly treated as an initial input to identify, describe, or disambiguate an entity, while subsequent evidence acquisition is largely driven by textual queries and webpage reading. 
This limits their ability to exploit visual evidence as an active part of the search trajectory. 
In realistic open-web search, visual observations such as logos, inscriptions, posters, emblems, screenshots, and spatial layouts can reveal new entities or constraints that determine what the agent should search next. 
Our work therefore focuses on evaluating whether multimodal search agents can not only retrieve and understand visual evidence, but also use it as a search pivot in interleaved language--vision search.

\subsection{Multimodal Agentic Search Benchmark }

The rapid progress of MLLMs has motivated benchmarks that move beyond static visual question answering toward open-world search, evidence gathering, and tool use. Early browsing benchmarks, such as BrowseComp~\citep{wei2025browsecomp}, primarily evaluate whether agents can perform difficult multi-hop web search and synthesize textual evidence, emphasizing browsing depth and final-answer correctness~\citep{zhang2026browsecomp}. Subsequent multimodal search benchmarks incorporate visual information into this process. For example, works like MMSearch and FVQA-Test extend search-based evaluation to multimodal inputs, requiring agents to reason over user-provided images together with external evidence~\citep{wu2025mmsearch,jiang2024mmsearch,li2025mm}. However, in these settings, visual information is largely pre-specified by the task, typically appearing as the initial query image or auxiliary context, rather than being actively sought by the agent during search.

More recent benchmarks further increase the visual complexity of multimodal browsing. BrowseComp-VL and VDR-Bench introduce richer visual inputs, region-level inspection, cropping, and noisy web environments~\citep{geng2026webwatcher,zeng2026vision}. These benchmarks make visual understanding more demanding, but they still mainly evaluate how agents interpret given or retrieved visual evidence, rather than whether agents can actively acquire new visual evidence as part of the search process. Recent visual browsing benchmarks, such as BrowseComp-$V^3$ and VisBrowse, take an important step by introducing active image search~\citep{zhang2026browsecomp,visbrowse}. Nevertheless, the retrieved visual evidence is often used as an endpoint for final VQA-style verification, such as reading a color, counting objects, or recognizing a person in the final image. As a result, these benchmarks underexplore the role of visual evidence as a search pivot that determines subsequent retrieval targets.

InterLV-Search differs by targeting interleaved multimodal search. Rather than treating images as given context or final-step evidence, our benchmark requires agents to actively acquire visual evidence, use it to guide later search, and repeatedly transition between textual and visual evidence. This setting better reflects open-web information seeking, where visual cues discovered during browsing can determine the next query, entity, page, tool call, or branch decision, and where agents must integrate evidence across longer and sometimes branching multimodal trajectories.

\section{Effect of Interaction Budget}
\begin{figure*}[h]
    \centering
    \includegraphics[width=0.95\textwidth]{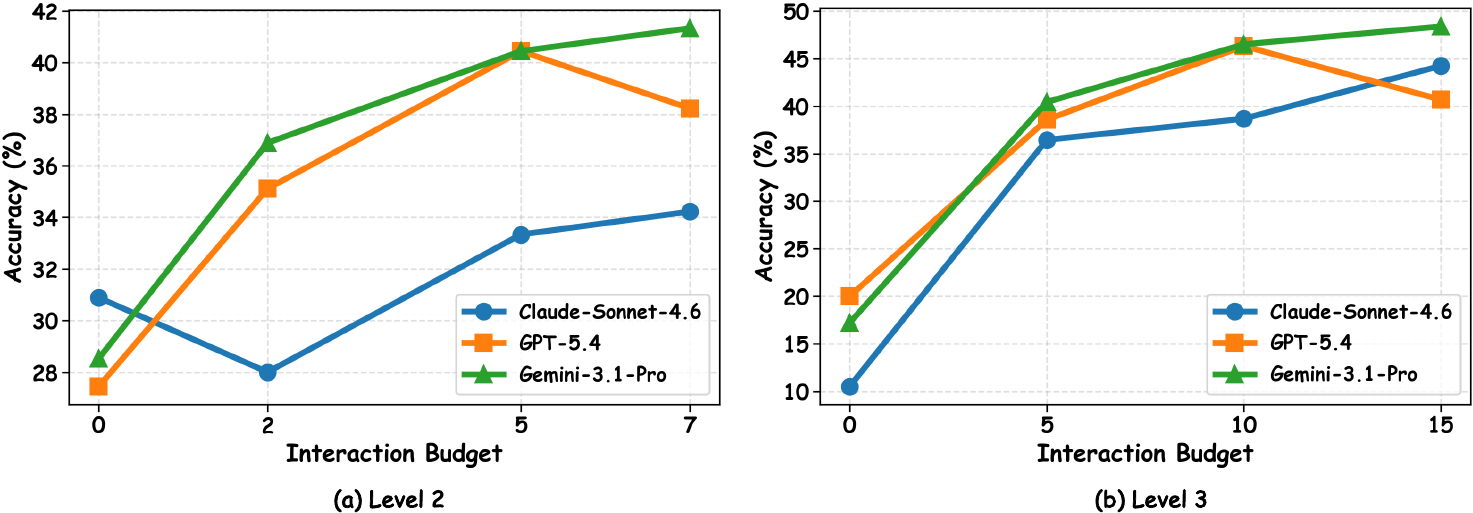}
    \caption{
    Effect of interaction budget on Level~2 and Level~3. 
    Level~2 uses smaller budgets because controlled offline chains are shorter, while Level~3 requires larger budgets for open-web search, branch exploration, and error recovery.
    }
    \label{fig:budget_ablation}
\end{figure*}

Figure~\ref{fig:budget_ablation} shows how model performance changes with different interaction budgets. On Level~2, accuracy improves when the budget increases from direct answering to a small number of tool interactions, but the gains quickly saturate around 5--7 interactions. This is consistent with the controlled offline setting: evidence paths are fixed, and the main challenge is whether the model can follow the intended evidence-to-query chain rather than repeatedly explore alternative sources.

On Level~3, the effect of budget is much stronger. Increasing the budget from direct answering to 5 interactions brings a large improvement for all models, showing that open-web interleaved search requires active evidence acquisition. Larger budgets further help models explore alternative webpages, recover from noisy evidence, and handle branch comparisons. However, the gains are not strictly monotonic for every model, suggesting that additional tool calls can also introduce distractors if the model cannot effectively filter and integrate retrieved evidence. Overall, the budget ablation supports that InterLV-Search evaluates long-horizon interleaved search rather than shallow single-step retrieval.

\section{Details of Agentic Framework}
\label{app:agent_details}
\begin{figure*}[t]
    \centering
    \includegraphics[width=0.95\textwidth]{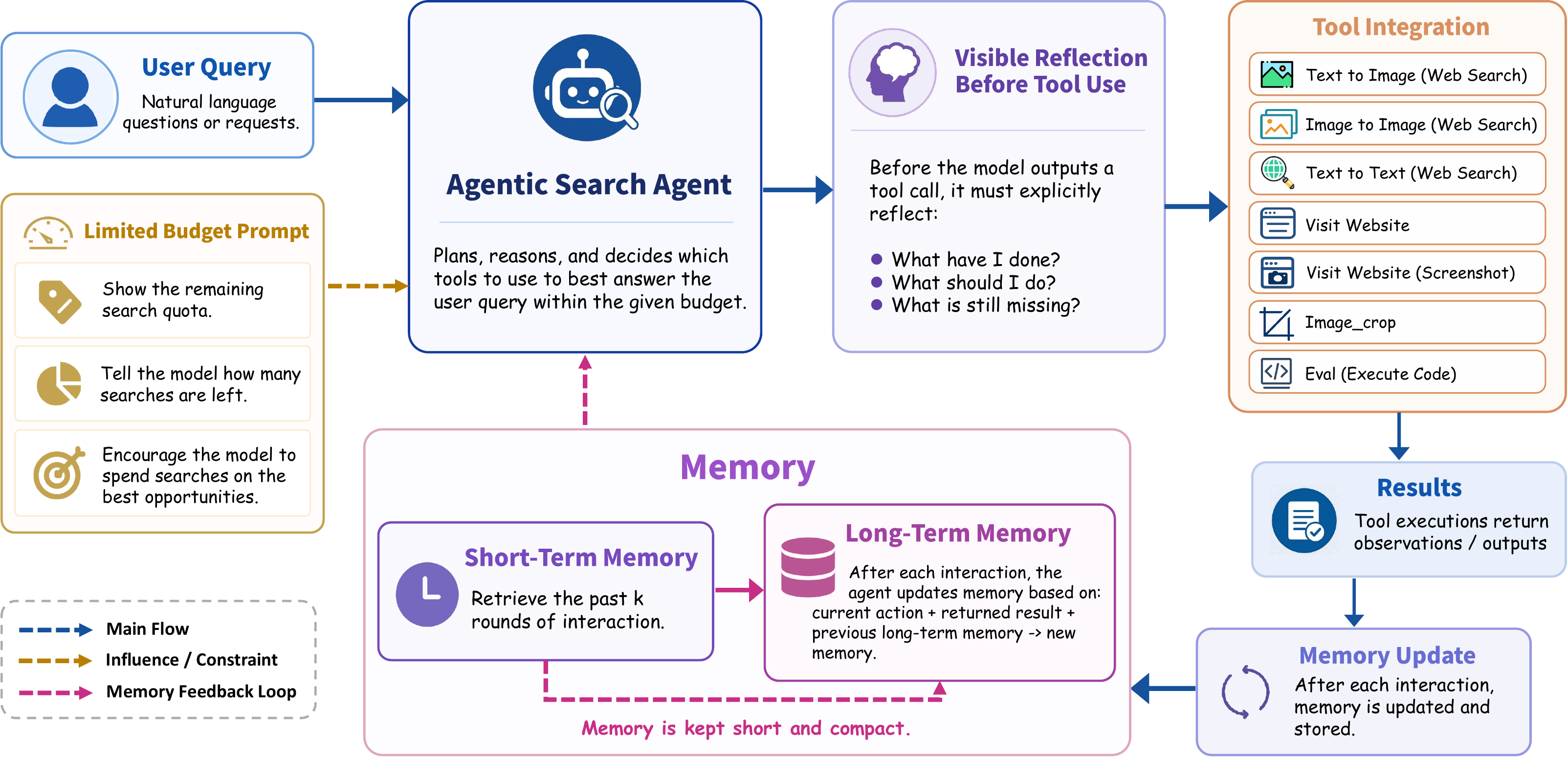
    }
    \caption{
    Overview of InterLV-Agent. The agent follows a reason-act-observe loop with limited interaction budgets, multimodal tool integration, and lightweight two-level memory. Short-term memory stores recent interactions, while long-term memory summarizes accumulated evidence and unresolved subgoals for subsequent search steps.
    }
    \label{fig:interlv_agent}
\end{figure*}

Figure~\ref{fig:interlv_agent} shows the InterLV-Agent workflow. The agent follows a reason-act-observe loop: it receives a user query, reflects on the current search state, selects a tool under a limited interaction budget, observes the returned result, and updates its memory before the next step. The framework supports multimodal tools such as text-to-image search, image-to-image search, web search, webpage browsing, screenshot browsing, image cropping, and code execution. A lightweight two-level memory stores recent interactions and compact long-term summaries, enabling standardized tool use, trajectory logging, and evaluation across models.
\subsection{Memory Implementation}

InterLV-Agent maintains a lightweight running memory to support long-horizon interleaved search. At each interaction step, the agent observes the previous memory, the current tool query proposed by the model, and the tool-returned result. These elements are passed to a memory-update prompt, which produces an updated running memory for the next step. Formally, the memory update takes the form:
\[
M_t = \mathrm{Update}(M_{t-1}, q_t, o_t),
\]
where $M_{t-1}$ is the previous running memory, $q_t$ is the current tool query, and $o_t$ is the returned observation.

We use a two-level memory design. The short-term memory directly stores the most recent interaction rounds in a formatted form, including tool names, tool queries, and observations. This gives the agent access to recent local context without additional summarization loss. The long-term memory compresses previous interactions into concise natural-language summaries that record key entities, retrieved evidence, visual clues, and unresolved subgoals. This design keeps the context compact while preserving the search state needed for long interleaved trajectories.

\subsection{Tool Implementation}
InterLV-Agent provides a unified tool interface for both online open-web search and offline controlled retrieval.

\paragraph{Online tools.}
For Level~3, we implement online tools based on external search and browser interfaces. The supported tools include:
\begin{itemize}
    \item \textbf{Image search}: given a textual query, returns relevant images together with their source URLs.
    \item \textbf{Web search}: given a textual query, returns webpage titles, snippets, and URLs.
    \item \textbf{Reverse image search}: given an input image, returns visually similar images and associated webpage information.
    \item \textbf{Webpage browsing}: opens a webpage and returns its textual content.
    \item \textbf{Screenshot browsing}: captures the current browser viewport as an image using Playwright.
    \item \textbf{Image cropping}: given an image and a bounding box, returns the cropped image region.
    \item \textbf{Code execution}: executes a model-generated code snippet and returns the output.
\end{itemize}
The web and image search interfaces are implemented with SerpAPI. Screenshot browsing is implemented with Playwright, allowing the agent to inspect webpage layouts and visual evidence when text-only page content is insufficient. We use top-5 search results as the observation of model.

\paragraph{Offline local retrieval tools.}
For Level~1 and Level~2, we use Qwen3-VL-Embedding-2B to build a local multimodal retrieval index over the benchmark corpus. This enables controlled evaluation without live-web stochasticity. The local tools support several retrieval modes:
\begin{itemize}
    \item \textbf{Local text search}: text query to textual entity information.
\begin{verbatim}
<query>{"skill": "local_text_search",
         "query": "...", "top_k": 5}</query>
\end{verbatim}

    \item \textbf{Local text search with image}: text query to entity items, returning both entity metadata and associated images.
\begin{verbatim}
<query>{"skill": "local_text_search_with_image",
         "query": "...", "top_k": 5}</query>
\end{verbatim}

    \item \textbf{Local text-to-image search}: text query to images, mainly used when the query describes visual appearance.
\begin{verbatim}
<query>{"skill": "local_text_to_image_search",
         "query": "...", "top_k": 5}</query>
\end{verbatim}

    \item \textbf{Local image search}: image query to visually similar images.
\begin{verbatim}
<query>{"skill": "local_image_search",
         "image": "img_1", "top_k": 5}</query>
\end{verbatim}
\end{itemize}

These local retrieval tools allow Level~1 and Level~2 to evaluate visual evidence seeking and controlled interleaved search under a fixed corpus. In contrast, the online tools in Level~3 evaluate the same agentic search loop under realistic open-web conditions.

\subsection{Prompt of the Agentic Framework}

\begin{promptbox}{System Prompt}
You are an agentic-search controller used for evaluation.\\[0.8em]
You are doing an Interleaved Multimodal Search Work. You need to use several tools, such as image\_search, web\_search and so on to solve the search problem. Your job is to solve a user query by iteratively deciding whether to:\\
1) search the web,\\
2) search images,\\
3) inspect a page,\\
4) crop an image,\\
5) run short Python code,\\
6) summarize gathered text,\\
or 7) finish with a final answer.\\[0.8em]
Rules:\\
- Be concise and tool-oriented.\\
- Prefer one useful action at a time.\\
- When you need a tool, emit one of the supported tags.\\
- When enough evidence has been collected, emit \texttt{<done>}final answer\texttt{</done>}.\\
- Before each search action, briefly reflect in 1-3 sentences: what you already know, what information is still missing, and why you need these searches. This helps you make more precise search decisions within limited steps. Keep your reflection concise.\\
- If prior observations already answer the question, directly emit \texttt{<done>}...\texttt{</done>}.\\
- If you need visual evidence from a webpage (layout, poster, image, design, text rendered inside an image), or if fetch\_webpage\_text did not provide enough useful evidence, use browse\_web\_page.\\
- If you already know the answer or can reason it out from existing knowledge, do so directly --- you do not have to search for everything.\\
- Search results may contain inaccurate or irrelevant information, especially when your query is broad. Critically evaluate each result against the question and only use information that is clearly relevant and reliable.\\[0.8em]
Supported action tags:\\
1) Search/query:\\
\texttt{<query>\{"skill": "web\_search", "query": "...", "num": 5\}</query>}\\
\texttt{<query>\{"skill": "image\_search", "query": "...", "num": 5\}</query>}\\
\texttt{<query>\{"skill": "lens\_search", "image\_url": "..."\}</query>}\\[0.8em]
2) Explicit tool:\\
\texttt{<tool name="fetch\_webpage\_text">\{"url": "https://..."\}</tool>}\\
\texttt{<tool name="browse\_web\_page">\{"url": "https://..."\}</tool>}\\
\texttt{<tool name="summarize\_text">\{"text": "..."\}</tool>}\\[0.8em]
3) Python execution:\\
\texttt{<code>}\\
\texttt{print(...)}\\
\texttt{</code>}\\[0.8em]
4) Image crop:\\
\texttt{<clip>\{"image": "https://...", "bbox": [x1, y1, x2, y2]\}</clip>} bbox uses normalized coordinates from 0 to 1000.\\[0.8em]
5) Final answer:\\
\texttt{<done>...</done>}\\[0.8em]
Output discipline:\\
- If acting, only emit the action block(s).\\
- If answering, only emit one \texttt{<done>...</done>} block.
\end{promptbox}

\begin{promptbox}{Input Prompt}
User question:\\
\texttt{\{query\}}\\[0.8em]
Available skills:\\
\texttt{\{skill\_descriptions\}}\\[0.8em]
\textit{[If running memory is not empty, include the following block:]}\\
Running memory (accumulated knowledge from previous searches):\\
\texttt{\{running\_memory\}}\\[0.8em]
Recent observations:\\
\texttt{\{trace\_text\}}\\[0.8em]
Decide the next best action. If the answer is ready, output \texttt{<done>...</done>}.
\end{promptbox}

\begin{promptbox}{Memory Prompt}
You are a memory compressor for an agentic search system.\\[0.8em]
Your job: given the original question, the previous memory, and new observations from this round, produce a REWRITTEN memory from scratch.\\[0.8em]
CRITICAL: This is a RUNNING SUMMARY, not a log. Every update REPLACES the previous memory entirely.\\
- Do NOT append new content to old memory.\\
- Do NOT preserve old wording just because it existed before.\\
- Rewrite the entire memory each time, keeping only what is currently most valuable.\\
- If new observations contradict or supersede old information, drop the old.\\[0.8em]
STRICT RULES:\\
1. Output ONLY the memory. No explanations, no markdown fences, no commentary.\\
2. Keep it SHORT: 5-12 lines. Never exceed 15 lines.\\
3. Only include facts and candidates backed by actual evidence from observations.\\
4. Drop anything speculative, unsupported, or no longer relevant.\\
5. Do NOT plan next actions — the main agent handles its own planning.\\
6. Compress ruthlessly: if 3 searches said the same thing, summarize in 1 sentence.\\
7. Be STATE-CENTRIC, not search-centric. Write what IS known, not what was searched.\\[0.8em]
OUTPUT FORMAT:\\
goal: \texttt{<one line: what we need to find>}\\
status: \texttt{<one sentence: current progress>}\\
blocking\_gap: \texttt{<one line: exactly where the reasoning chain is stuck and what specific piece of information is missing to move forward>}\\[0.8em]
confirmed\_facts:\\
- \texttt{<ONLY positively confirmed facts with evidence — e.g. "X is Y", "X appeared in Z">}\\
- Do NOT write "we searched but didn't find" or "no evidence for X" here.\\
- Do NOT write unconfirmed candidates or speculations here.\\
- If nothing is confirmed yet, omit this field entirely.\\[0.8em]
best\_candidates:\\
- \texttt{<entity>}: \texttt{<concrete evidence supporting it>}\\[0.8em]
dead\_ends: \texttt{<one line summarizing what didn't work>}\\[0.8em]
key\_images:\\
- \texttt{<image\_id>}: \texttt{<what it shows>}, only give image description (what is it), include only images likely useful for future verification; do not list images unless they matter.\\[0.8em]
FIELD RULES:\\
- Omit any field with nothing to write.\\
- confirmed\_facts: STRICTLY positive confirmed facts only. "We didn't find X" is NOT a confirmed fact. Unconfirmed candidates do NOT belong here.\\
- best\_candidates: every entry MUST have evidence. No "weak/unsupported" entries.\\
- blocking\_gap: this is the most important field — clearly state what single piece of information, if found, would unblock the next step.\\
- dead\_ends: one line max, summarize patterns not individual queries.\\
- The entire output must read as a fresh, standalone state snapshot — not a search history. \\[0.8em]

---\\[0.8em]
\texttt{\{RUNNING\_MEMORY\_SYSTEM\_PROMPT\}}\\
Original question:\\
\texttt{\{query\}}\\[0.8em]
Current running memory:\\
\texttt{\{old\_memory\}} \textit{[If empty, output: "(empty --- this is the first round)"]}\\[0.8em]
New observations from this round:\\
\texttt{\{new\_observations\_text\}}\\[0.8em]
Now produce the updated running memory.
\end{promptbox}

\begin{promptbox}{GPT Judge Prompt}
You are an answer-equivalence judge.\\[0.8em]
You are given:\\
- QUESTION\\
- GOLD\_ANSWER\\
- PREDICTED\_ANSWER\\[0.8em]
IMPORTANT: The PREDICTED\_ANSWER field may contain the model's internal reasoning or chain-of-thought process before the actual answer. You MUST ignore any reasoning, thinking steps, or intermediate analysis. Focus ONLY on the final answer stated by the model --- the conclusion after all reasoning.\\[0.8em]
Judge whether the final answer in PREDICTED\_ANSWER should count as correct for this QUESTION, using GOLD\_ANSWER as the reference.\\[0.8em]
Guidelines:\\
- Use the QUESTION only to understand answer context and granularity.\\
- Do NOT use outside knowledge.\\
- Do NOT invent hidden requirements not supported by QUESTION, GOLD\_ANSWER, and PREDICTED\_ANSWER.\\
- Be reasonably permissive.\\
- Ignore all reasoning steps, thinking processes, or intermediate conclusions --- evaluate only the final answer.\\[0.8em]
Return YES if the predicted final answer is clearly the same answer as the gold answer in context, including:\\
- same answer in different wording\\
- alias / synonym / alternative surface form\\
- numeral vs word form\\
- full sentence containing the answer\\
- a more specific form of the same answer\\
- harmless extra descriptive words\\[0.8em]
Return NO if the predicted final answer is clearly a different answer.\\[0.8em]
Examples:\\
QUESTION: What Roman numeral marks the day on the tablet?\\
GOLD\_ANSWER: IV\\
PREDICTED\_ANSWER: The day is marked by IV.\\
YES\\[0.8em]
QUESTION: How many helmets appear on that reverse?\\
GOLD\_ANSWER: Two\\
PREDICTED\_ANSWER: Two women.\\
YES\\[0.8em]
QUESTION: What object is in its talons?\\
GOLD\_ANSWER: A boomerang.\\
PREDICTED\_ANSWER: Each kangaroo holds a boomerang.\\
YES\\[0.8em]
QUESTION: What is the floral emblem?\\
GOLD\_ANSWER: A rose.\\
PREDICTED\_ANSWER: The floral emblem is the White Rose of York.\\
YES\\[0.8em]
QUESTION: What year?\\
GOLD\_ANSWER: 1970\\
PREDICTED\_ANSWER: 1990\\
NO\\[0.8em]
Output exactly one token:\\
YES\\
or\\
NO\\[0.8em]
QUESTION: \texttt{\{question\}}\\
GOLD\_ANSWER: \texttt{\{gold\}}\\
PREDICTED\_ANSWER: \texttt{\{pred\}}
\end{promptbox}

\section{Case Study}
\begin{figure*}[h]
    \centering
    \includegraphics[width=0.8\textwidth]{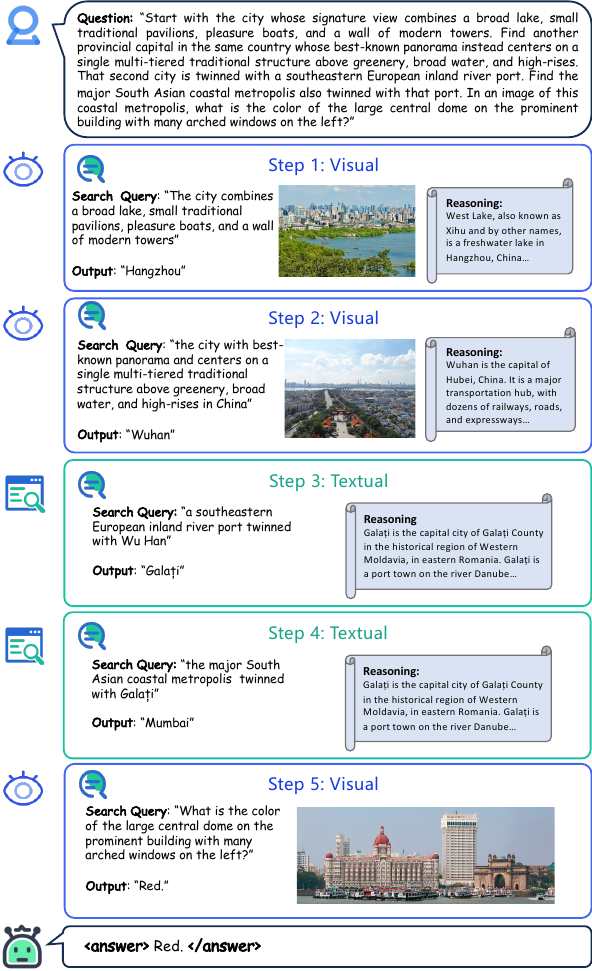}
    \caption{
    A Case of Level 2.
    }
    \label{fig:level2_case}
\end{figure*}

\begin{figure*}[t]
    \centering
    \includegraphics[width=0.6\textwidth]{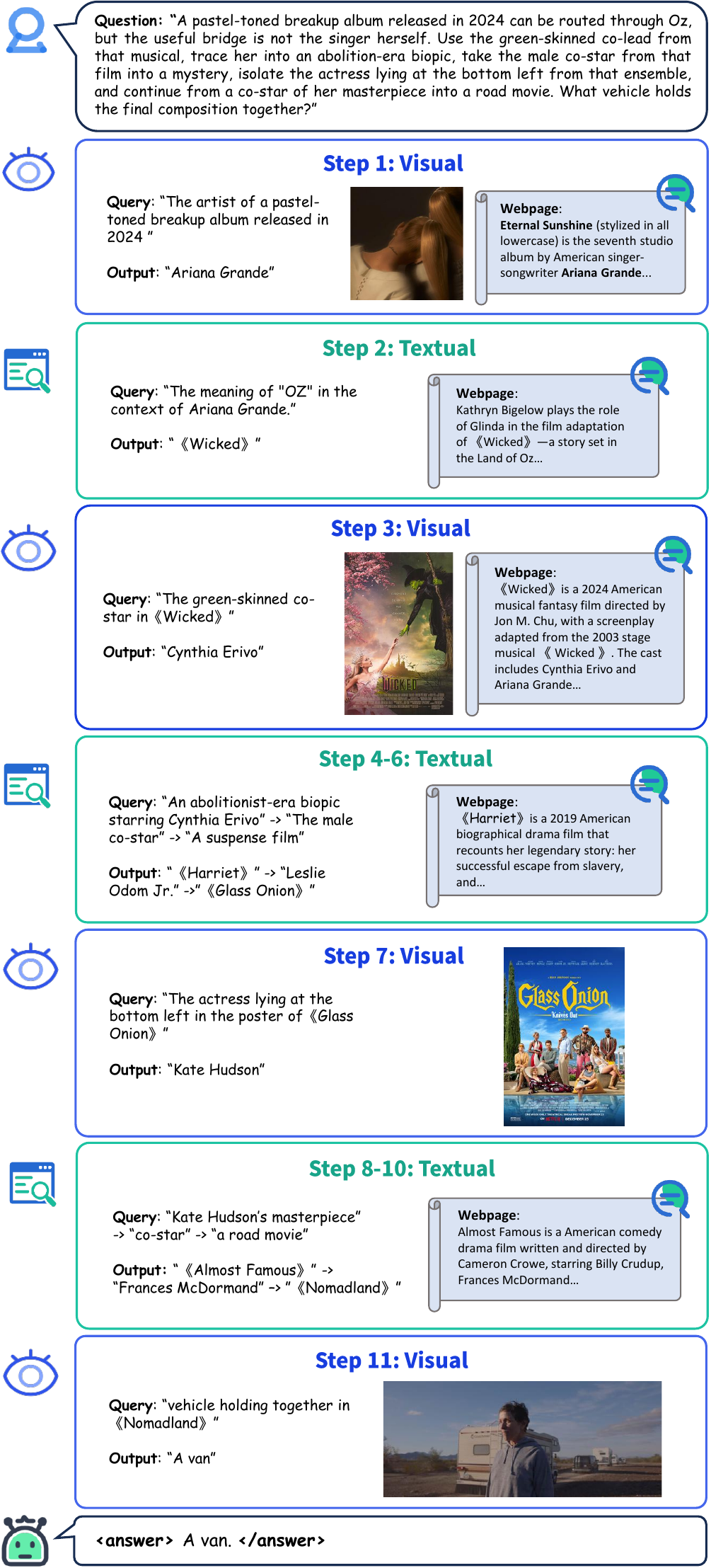}
    \caption{
    A Single-Chain Case of Level 3.
    }
    \label{fig:level3_case1}
\end{figure*}

\begin{figure*}[t]
    \centering
    \includegraphics[width=0.95\textwidth]{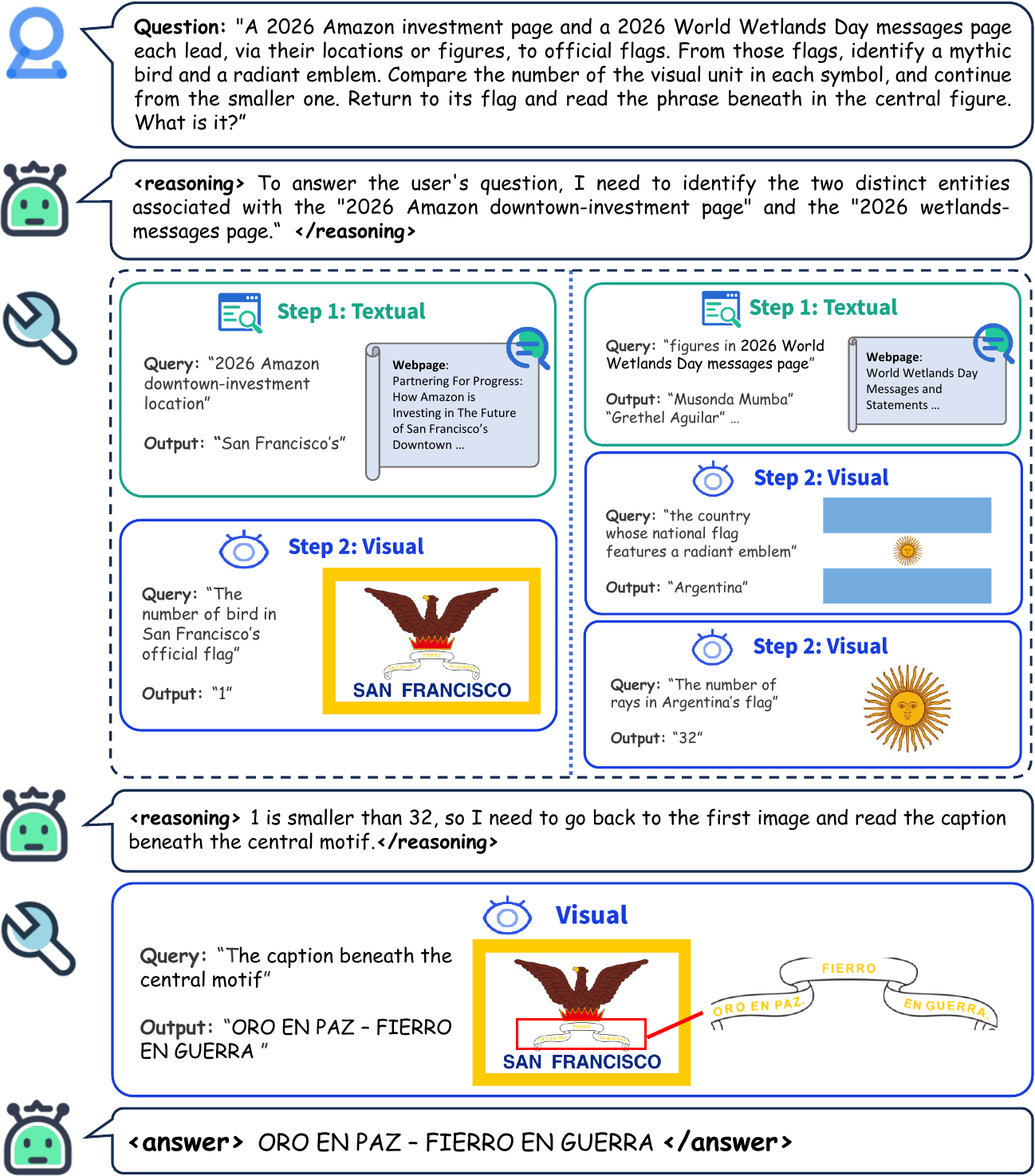}
    \caption{
    A Multi-Branch Case of Level 3.
    }
    \label{fig:level3_case2}
\end{figure*}

\FloatBarrier 
\label{app:case_study}

\subsection{Level 1 Case}
Figure~\ref{fig:teaser_compare} and Figure~\ref{fig:case_study} present two Level~1 cases of InterLV-Search. In this type of task, the question first provides a textual information need that requires reasoning over semantic clues to infer what visual evidence should be sought. The model must then actively retrieve the target image from the local image collection and inspect whether the retrieved visual evidence matches the question context. After locating the relevant image, the model further answers a fine-grained visual question based on image details. This setting evaluates the primitive capability of active visual evidence seeking: the agent must reason from text to decide what to search for, acquire the corresponding visual evidence, and ground the final answer in that evidence.

\subsection{Level 2 Case}

The Level~2 case in Fig.~\ref{fig:teaser_compare} illustrates how controlled interleaved search uses visual evidence as intermediate search pivots rather than only as the final VQA source. The trajectory contains multiple visual search blocks that ground intermediate entities, such as cities identified through distinctive skyline or landmark descriptions, interwoven with textual relation blocks such as sister-city and country constraints. The search then returns to a terminal image for fine-grained VQA. Another Level~2 example shown in Fig.~\ref{fig:case_study} is presented in greater detail in Fig.~\ref{fig:level2_case}, which shows the complete interleaved Multimodal search trajectory.

\subsection{Level 3 Case}

Figure~\ref{fig:teaser_compare} illustrates a multi-branch Level~3 case in the open web, where the agent must explore several visually grounded routes in parallel, compare textual evidence across branches, and continue from the selected route. The task begins with multiple visual clues tied to different film pages, requiring the agent to localize candidate images or webpages, verify them with textual metadata such as title, year, and runtime, and then use the comparison result to decide which branch survives. The final answer is obtained only after returning to the selected branch and inspecting the target visual evidence. This case demonstrates how Level~3 evaluates not only open-web search, but also branch-level search control and multimodal evidence integration.

Besides the multi-branch case, Fig.~\ref{fig:level3_case1} shows a single-chain Level~3 example with a long interleaved trajectory. The question starts from a visually described music-related clue and repeatedly moves through film posters, cast members, and film pages. The agent first uses visual search to identify the artist associated with the pastel-toned breakup album, then uses textual evidence to route through \textit{Wicked}. The poster and visual appearance of the green-skinned co-lead become the next search pivot, leading to Cynthia Erivo and then to \textit{Harriet}. Subsequent textual steps follow co-star and film-role relations, but the chain repeatedly returns to visual evidence, such as the \textit{Glass Onion} ensemble poster and the final road-movie composition. Thus, the example is not a simple text-only movie chain: visual evidence repeatedly determines which person, film, or poster should be searched next. The final answer requires inspecting the terminal image to identify the vehicle that holds the composition together. We also show another case of multi-branch in figure~\ref{fig:level3_case2}

Together, these cases show the two main Level~3 patterns in InterLV-Search. Multi-branch examples test whether agents can search and compare parallel routes before continuing, while single-chain examples test whether agents can maintain a long open-web trajectory in which visual evidence repeatedly acts as a search pivot. Both settings require agents to alternate between visual localization and textual verification, rather than treating images as either given inputs or final VQA endpoints.

\subsection{Success and Failure Cases on Level 3}

Level~3 is the most realistic setting in InterLV-Search, requiring open-web search, long search-state maintenance, and branch comparison. We show one successful and one failed Example below to illustrate why interleaved multimodal search requires visual evidence as a search pivot rather than ordinary text-only browsing.

The successful case grounds three visually described film branches, compares their textual runtimes, and returns to the selected branch for terminal visual inspection. In contrast, the multi-branch failed case mostly uses broad textual search and page fetching; it never visually grounds the Tate/Picasso and Berlinale/Niigata branches into their corresponding local-symbol systems, and therefore fails to obtain the branch counts or inspect the final border color. These cases highlight the core challenge targeted by InterLV-Search: agents must use visual evidence as intermediate search pivots, preserve multi-branch search state, compare multimodal evidence across routes, and continue searching from the selected branch. This is precisely the capability that endpoint-oriented visual browsing or text-centric web search fails to isolate. 
% Requires:
% \usepackage[most]{tcolorbox}
% \usepackage{xcolor}
% \usepackage{enumitem}
% \usepackage{hyperref}
%
% Optional:
% \setlist[itemize]{leftmargin=1.2em,itemsep=2pt,topsep=2pt}
% \setlist[enumerate]{leftmargin=1.6em,itemsep=2pt,topsep=2pt}

% =========================================================
% Case 1: Correct Example
% =========================================================

\begin{trajbox}{Case 1: Successed Example}{green!60!black}

\stepbar{Question}
\small
A pale-haired rear-view poster, an orange-black volcanic title slab, and a pink handwritten title on sky-blue each lock a different film page. Follow the three lines into official festival pages, compare runtimes, keep the middle result, and continue. On the surviving branch's target image, what natural event rises behind the title?

\stepbar{Step 1}
\tagthink

This is a three-branch visual search question. Each branch starts from a poster-like visual clue, so I should first locate candidate film pages by searching the visual descriptions.

\stoptagthink

\vspace{0.3em}
\tagtoolcall
\begin{traceblock}
\textless query\textgreater\{"skill": "web\_search", "query": "\textbackslash"pink handwritten title on sky-blue\textbackslash" film poster", "num": 5\}\textless/query\textgreater\\
\textless query\textgreater\{"skill": "web\_search", "query": "\textbackslash"orange-black volcanic title slab\textbackslash" film poster", "num": 5\}\textless/query\textgreater
\end{traceblock}
\stoptagtoolcall

\vspace{0.3em}
\tagtoolresponse

The first broad search is noisy:
\begin{enumerate}
  \item \textbf{Caption:} Pink movie posters - IMDb.\\
  \textbf{Snippet:} A broad list of pink poster designs; not enough to identify the specific festival page.
  \item \textbf{Caption:} Pink and Blue Movie Poster Design - Pinterest.\\
  \textbf{Snippet:} Decorative poster material; visually related but not the official film page.
  \item \textbf{Caption:} Pretty Pink Posters - Letterboxd.\\
  \textbf{Snippet:} A large user list; not a reliable terminal source.
  \item \textit{\ldots}
\end{enumerate}

\stoptagtoolresponse

\stepbar{Step 2}
\tagthink

The textual search is too broad. Because the branch clues are visual, I switch to image search for the poster descriptions.

\stoptagthink

\vspace{0.3em}
\tagtoolcall
\begin{traceblock}
\textless query\textgreater\{"skill": "image\_search", "query": "\textbackslash"pale-haired rear-view\textbackslash" film poster", "num": 5\}\textless/query\textgreater\\
\textless query\textgreater\{"skill": "image\_search", "query": "\textbackslash"orange-black volcanic title slab\textbackslash" film poster", "num": 5\}\textless/query\textgreater
\end{traceblock}
\stoptagtoolcall

\vspace{0.3em}
\tagtoolresponse

The image search now exposes candidate visual pages:
\begin{enumerate}
  \item \textbf{Caption:} rear-view / pale-haired film poster results.\\
  \textbf{Observation:} candidates visually match the ``pale-haired rear-view'' branch.
  \item \textbf{Caption:} volcanic orange-black poster results.\\
  \textbf{Observation:} candidates include volcanic poster imagery, but some hits are older or generic.
  \item \textbf{Caption:} Volcano movie poster / volcanic title imagery.\\
  \textbf{Observation:} useful visual motif, but must be checked against the 2025--2026 festival-page constraint.
  \item \textit{\ldots}
\end{enumerate}

\stoptagtoolresponse

\stepbar{Step 3}
\tagthink

The image results give visual candidates, but I need official festival pages and runtimes. I follow the three visual lines into their text records.

\stoptagthink

\vspace{0.3em}
\tagtoolcall
\begin{traceblock}
\textless query\textgreater\{"skill": "web\_search", "query": "\textbackslash"Nox\textbackslash" film festival runtime pale-haired rear-view poster", "num": 5\}\textless/query\textgreater\\
\textless query\textgreater\{"skill": "web\_search", "query": "\textbackslash"Krakatoa\textbackslash" film festival runtime orange black volcanic poster", "num": 5\}\textless/query\textgreater\\
\textless query\textgreater\{"skill": "web\_search", "query": "\textbackslash"My Semba\textbackslash" film festival runtime pink handwritten title sky blue", "num": 5\}\textless/query\textgreater
\end{traceblock}
\stoptagtoolcall

\vspace{0.3em}
\tagtoolresponse

The official/festival-page route resolves the three branches:
\begin{enumerate}
  \item \textbf{Caption:} Nox - festival / film page.\\
  \textbf{Snippet:} Runtime listed as 64 minutes.
  \item \textbf{Caption:} Krakatoa - festival / film page.\\
  \textbf{Snippet:} Runtime listed as 79 minutes.
  \item \textbf{Caption:} My Semba - festival / film page.\\
  \textbf{Snippet:} Runtime listed as 93 minutes.
  \item \textit{\ldots}
\end{enumerate}

\stoptagtoolresponse

\stepbar{Step 4}
\tagthink

Now I can compare the three text values recovered from visually-locked branches:
Nox = 64, Krakatoa = 79, My Semba = 93. The middle runtime is 79, so the surviving branch is Krakatoa.

\stoptagthink

\vspace{0.3em}
\tagtoolcall
\begin{traceblock}
 compare 64, 79, 93  --> keep middle branch = Krakatoa 
\end{traceblock}
\stoptagtoolcall

\vspace{0.3em}
\tagtoolresponse

The branch comparison selects:
\begin{enumerate}
  \item \textbf{Winning branch:} Krakatoa.
  \item \textbf{Reason:} 79 minutes is the median runtime.
  \item \textbf{Next action:} return to the Krakatoa target image rather than the other two posters.
  \item \textit{\ldots}
\end{enumerate}

\stoptagtoolresponse

\stepbar{Step 5}
\tagthink

The question asks about the surviving branch's target image, not merely the title. I search visually for the Krakatoa target image and ignore the letterforms.

\stoptagthink

\vspace{0.3em}
\tagtoolcall
\begin{traceblock}
\textless query\textgreater\{"skill": "image\_search", "query": "\textbackslash"Krakatoa\textbackslash" film poster volcanic eruption title", "num": 5\}\textless/query\textgreater
\end{traceblock}
\stoptagtoolcall

\vspace{0.3em}
\tagtoolresponse

The terminal image route shows:
\begin{enumerate}
  \item \textbf{Caption:} Krakatoa poster / target image.\\
  \textbf{Observation:} orange-black title slab with a dramatic natural event behind the typography.
  \item \textbf{Caption:} Krakatoa festival image.\\
  \textbf{Observation:} the relevant background is not a static mountain or cloud bank.
  \item \textbf{Caption:} related volcanic imagery.\\
  \textbf{Observation:} useful for checking that the background event is an eruption.
  \item \textit{\ldots}
\end{enumerate}

\stoptagtoolresponse

\stepbar{Step 6}
\tagthink

I inspect the target image on the winning branch. The task says to ignore the letterforms and read the natural event rising behind the title.

\stoptagthink

\vspace{0.3em}
\tagtoolcall
\begin{traceblock}
\textless visual\_inspection\textgreater surviving image = Krakatoa target image; ignore typography; inspect background event \textless/visual\_inspection\textgreater
\end{traceblock}
\stoptagtoolcall

\vspace{0.3em}
\tagtoolresponse

The background behind the title is a volcanic plume / eruption rising upward.

\stoptagtoolresponse

\stepbar{Step 7}
\tagthink

The answer is therefore the natural event visible behind the title on the selected Krakatoa branch.

\stoptagthink

\vspace{0.3em}
\taganswer
\begin{traceblock}
Volcanic eruption
\end{traceblock}
\stoptaganswer

\vspace{0.4em}
\textbf{Final Answer:} \textbf{\textcolor{green!60!black}{A volcanic eruption}}

\stepbar{Why this succeeds}
\begin{itemize}
  \setlength{\itemsep}{2pt}
  \item The model must first use image search to lock three different film branches.
  \item It then switches to text access to read runtimes from official/festival pages.
  \item It performs a branch comparison and only then returns to the winning branch's terminal image.
  \item The final VQA is not a title lookup; it requires visual inspection of the background behind the title.
\end{itemize}

\end{trajbox}

%\newpage

% =========================================================
% Case 2: Failed Example
% =========================================================

\begin{trajbox}{Case 2: Failed Multi-Branch Example}{red!75!black}

\stepbar{Question}
\small
A Tate page centred on Picasso and a Berlinale page set in Niigata each open separate routes into local symbol systems. Continue from the line with the smaller count. What colour occupies the left half of the border on that winning branch?

\stepbar{Step 1}
\tagthink

I start by searching the whole question phrase, hoping that the Tate/Picasso
and Berlinale/Niigata clues will co-occur.

\stoptagthink

\vspace{0.3em}
\tagtoolcall
\begin{traceblock}
\textless query\textgreater\{"skill": "web\_search", "query": "Tate page centred on Picasso Berlinale page set in Niigata", "num": 5\}\textless/query\textgreater
\end{traceblock}
\stoptagtoolcall

\vspace{0.3em}
\tagtoolresponse

A web search found:
\begin{enumerate}
  \item \textbf{Caption:} Theatre Picasso \textbar{} Tate Modern.\\
  \textbf{Snippet:} This exhibition, staged by contemporary artists, sheds new light on Picasso's work.
  \item \textbf{Caption:} A Staged Re-Examination of Picasso's Performative Genius at Tate.\\
  \textbf{Snippet:} The installation marks the centenary of \textit{The Three Dancers}.
  \item \textbf{Caption:} `Theatre Picasso' At Tate Modern Is A Tribute To `The Three Dancers'.\\
  \textbf{Snippet:} At the centre is \textit{The Three Dancers}.
  \item \textit{\ldots}
\end{enumerate}

\stoptagtoolresponse

\stepbar{Step 2}
\tagthink

The Tate side is visible, but the task is a multi-branch comparison. I now try
to find the Niigata-side source.

\stoptagthink

\vspace{0.3em}
\tagtoolcall
\begin{traceblock}
\textless query\textgreater\{"skill": "web\_search", "query": "\textbackslash"Berlinale\textbackslash" \textbackslash"Niigata\textbackslash"", "num": 5\}\textless/query\textgreater
\end{traceblock}
\stoptagtoolcall

\vspace{0.3em}
\tagtoolresponse

The search finds the Niigata-side page:
\begin{enumerate}
  \item \textbf{Caption:} Shibire \textbar{} Numb - Berlinale.\\
  \textbf{Snippet:} Born in 1992 in Niigata, Japan; studied at Bunka Fashion College before turning to film.
  \item \textbf{Caption:} Berlinale 2026: Numb interview with Takuya Uchiyama.\\
  \textbf{Snippet:} Between Niigata and Tokyo, going to Tokyo is seen as positive.
  \item \textbf{Caption:} NUMB - Berlinale PDF.\\
  \textbf{Snippet:} A story of belonging and identity in the frozen winter of Niigata.
  \item \textit{\ldots}
\end{enumerate}

\stoptagtoolresponse

\stepbar{Step 3}
\tagthink

Instead of grounding each branch into an official emblem or symbol page, I try
to search the abstract phrase ``local symbol systems''. This is the first drift.

\stoptagthink

\vspace{0.3em}
\tagtoolcall
\begin{traceblock}
\textless query\textgreater\{"skill": "web\_search", "query": "\textbackslash"local symbol systems\textbackslash"", "num": 5\}\textless/query\textgreater
\end{traceblock}
\stoptagtoolcall

\vspace{0.3em}
\tagtoolresponse

The results are generic and not branch-grounded:
\begin{enumerate}
  \item \textbf{Caption:} Modern disruption of traditional symbol systems.\\
  \textbf{Snippet:} Discusses local symbol systems in traditional communities.
  \item \textbf{Caption:} Did the illiterate lower classes of ancient times create any symbols?\\
  \textbf{Snippet:} Mentions local symbol systems but not civic emblems.
  \item \textbf{Caption:} Understanding Nsibidi Symbols and History.\\
  \textbf{Snippet:} Discusses symbol systems, not Málaga or Niigata.
  \item \textit{\ldots}
\end{enumerate}

\stoptagtoolresponse

\stepbar{Step 4}
\tagthink

I continue with broad variants rather than completing the two visual branches.
These searches retrieve related pages but do not produce the needed counts.

\stoptagthink

\vspace{0.3em}
\tagtoolcall
\begin{traceblock}
\textless query\textgreater\{"skill": "web\_search", "query": "\textbackslash"Berlinale\textbackslash" \textbackslash"Niigata\textbackslash" site:en.wikipedia.org", "num": 5\}\textless/query\textgreater\\
\textless query\textgreater\{"skill": "web\_search", "query": "\textbackslash"Tate\textbackslash" \textbackslash"Picasso\textbackslash" \textbackslash"local symbol systems\textbackslash"", "num": 5\}\textless/query\textgreater
\end{traceblock}
\stoptagtoolcall

\vspace{0.3em}
\tagtoolresponse

The search drifts:
\begin{enumerate}
  \item \textbf{Caption:} Yasuhiro Morinaga - Wikipedia.\\
  \textbf{Snippet:} Mentions Niigata City Art Museum and Berlinale history, not the symbol count.
  \item \textbf{Caption:} Theatre Picasso \textbar{} Tate Modern.\\
  \textbf{Snippet:} Picasso exhibition page, not the Málaga emblem branch.
  \item \textbf{Caption:} Tate Modern's Theatre Picasso review.\\
  \textbf{Snippet:} Review context; no civic-symbol route.
  \item \textit{\ldots}
\end{enumerate}

\stoptagtoolresponse

\stepbar{Step 5}
\tagthink

I fetch the Tate and Berlinale pages, but fetching the source pages is not enough:
the missing action is to route from them into the two local symbol systems and
compare the counts.

\stoptagthink

\vspace{0.3em}
\tagtoolcall
\begin{traceblock}
\textless tool name="fetch\_webpage\_text"\textgreater\{"url": "https://www.tate.org.uk/whats-on/tate-modern/theatre-picasso"\}\textless/tool\textgreater\\
\textless tool name="fetch\_webpage\_text"\textgreater\{"url": "https://www.berlinale.de/en/2026/programme/202614502.html"\}\textless/tool\textgreater
\end{traceblock}
\stoptagtoolcall

\vspace{0.3em}
\tagtoolresponse

The visited pages give only the entry contexts:
\begin{enumerate}
  \item \textbf{Fetched page title:} Theatre Picasso \textbar{} Tate Modern.\\
  \textbf{Body excerpt:} Picasso exhibition; \textit{The Three Dancers}; Tate collection.
  \item \textbf{Fetched page title:} Shibire \textbar{} Numb \textbar{} Berlinale.\\
  \textbf{Body:} empty or not useful for branch comparison.
  \item \textit{\ldots}
\end{enumerate}

\stoptagtoolresponse

\stepbar{Step 6}
\tagthink

Because I never grounded the two branches, I never count Málaga's two saint
figures against Niigata's seven official symbols. I also never inspect the
winning branch's border.

\stoptagthink

\vspace{0.3em}
\taganswer
\begin{traceblock}
blue
\end{traceblock}
\stoptaganswer

\vspace{0.4em}
\textbf{Model Answer:} \textbf{\textcolor{red!75!black}{Blue}}\\
\textbf{Correct Answer:} \textbf{\textcolor{green!60!black}{Purple}}

\stepbar{Expected reasoning}
\begin{itemize}
  \setlength{\itemsep}{2pt}
  \item Picasso/Tate branch $\rightarrow$ Málaga emblem $\rightarrow$ count two saints.
  \item Berlinale/Niigata branch $\rightarrow$ Niigata official symbols $\rightarrow$ count seven symbols.
  \item Compare $2$ and $7$; keep the smaller branch, Málaga.
  \item On the Málaga emblem, ignore the saints, fortress, and waves; inspect the border colours.
  \item The left half of the border is purple.
\end{itemize}

\stepbar{Why this fails}
\begin{itemize}
  \setlength{\itemsep}{2pt}
  \item The model searches hard, but mostly with broad textual queries.
  \item It never completes both visual branches before comparing.
  \item It does not inspect the terminal visual detail on the winning branch.
  \item The failure is not lack of search effort; it is failure to use visual branch structure.
\end{itemize}

\end{trajbox}

% =========================================================
% Case 3: Failed Single-Chain Example
% =========================================================

\begin{trajbox}{Case 3: Failed Single-Chain Example}{red!75!black}

\stepbar{Question}
\small
A summit-tourism page about double-decker buses becomes useful only if you treat the vehicles as a false endpoint and return to the guardian identity they borrow from. Once the state branch behind the host city is recovered, ignore the star and ask the diagonal question on the final flag instead. What colour is the band?

\stepbar{Step 1}
\tagthink

I start by searching the phrase about double-decker buses and guardian identity, hoping that the page and the borrowed visual identity will co-occur.

\stoptagthink

\vspace{0.3em}
\tagtoolcall
\begin{traceblock}
\textless query\textgreater\{"skill": "web\_search", "query": "\textbackslash"double-decker buses\textbackslash" \textbackslash"guardian identity\textbackslash"", "num": 5\}\textless/query\textgreater
\end{traceblock}
\stoptagtoolcall

\vspace{0.3em}
\tagtoolresponse

The search results are generic and do not ground the intended event route:
\begin{enumerate}
  \item \textbf{Caption:} Bus travel: Frequently asked questions.\\
  \textbf{Snippet:} Mentions guardian identity in an unrelated travel-policy context.
  \item \textbf{Caption:} Double-decker bus travel pages.\\
  \textbf{Snippet:} General bus-tourism information, not the target summit page.
  \item \textit{\ldots}
\end{enumerate}

\stoptagtoolresponse

\stepbar{Step 2}
\tagthink

The first query is too abstract. I search for summit pages involving double-decker buses, but I still stay in broad web-search mode rather than grounding the bus visual.

\stoptagthink

\vspace{0.3em}
\tagtoolcall
\begin{traceblock}
\textless query\textgreater\{"skill": "web\_search", "query": "\textbackslash"double-decker buses\textbackslash" \textbackslash"summit\textbackslash"", "num": 5\}\textless/query\textgreater
\end{traceblock}
\stoptagtoolcall

\vspace{0.3em}
\tagtoolresponse

The search remains noisy:
\begin{enumerate}
  \item \textbf{Caption:} Experiential Marketing Trend of the Week: Double-Decker Buses.\\
  \textbf{Snippet:} A broad marketing article; not the target climate or summit page.
  \item \textbf{Caption:} Short Stay Summit 2026.\\
  \textbf{Snippet:} Mentions red double-decker buses, but in an unrelated tourism context.
  \item \textbf{Caption:} Tourism summit pages.\\
  \textbf{Snippet:} Related to tourism, but not the intended host-city or guardian route.
  \item \textit{\ldots}
\end{enumerate}

\stoptagtoolresponse

\stepbar{Step 3}
\tagthink

I continue searching text variants of ``summit tourism'' and ``guardian identity''. This drifts away from the visual clue on the buses.

\stoptagthink

\vspace{0.3em}
\tagtoolcall
\begin{traceblock}
\textless query\textgreater\{"skill": "web\_search", "query": "\textbackslash"summit-tourism\textbackslash"", "num": 5\}\textless/query\textgreater\\
\textless query\textgreater\{"skill": "web\_search", "query": "\textbackslash"guardian identity\textbackslash" \textbackslash"false endpoint\textbackslash"", "num": 5\}\textless/query\textgreater
\end{traceblock}
\stoptagtoolcall

\vspace{0.3em}
\tagtoolresponse

The results are still not branch-grounded:
\begin{enumerate}
  \item \textbf{Caption:} Lee's Summit Tourism.\\
  \textbf{Snippet:} A tourism page for a place named Lee's Summit, unrelated to the target.
  \item \textbf{Caption:} Summit of the Americas tourism material.\\
  \textbf{Snippet:} Tourism-related, but not tied to the bus visual identity.
  \item \textbf{Caption:} Generic identity-guardian results.\\
  \textbf{Snippet:} Software or access-control pages, unrelated to the visual clue.
  \item \textit{\ldots}
\end{enumerate}

\stoptagtoolresponse

\stepbar{Step 4}
\tagthink

I search harder for double-decker buses and summit tourism, but I still do not use image search or visual inspection to identify the COP30 guardian identity on the bus graphics.

\stoptagthink

\vspace{0.3em}
\tagtoolcall
\begin{traceblock}
\textless query\textgreater\{"skill": "web\_search", "query": "\textbackslash"summit-tourism\textbackslash" \textbackslash"double-decker\textbackslash"", "num": 5\}\textless/query\textgreater\\
\textless query\textgreater\{"skill": "web\_search", "query": "\textbackslash"summit\textbackslash" \textbackslash"tourism\textbackslash" \textbackslash"double-decker buses\textbackslash"", "num": 10\}\textless/query\textgreater
\end{traceblock}
\stoptagtoolcall

\vspace{0.3em}
\tagtoolresponse

The search retrieves broad or unrelated pages about tourism, buses, and summits. None of them completes the intended route from the bus visual identity to the COP30 host city and then to the state flag.

\stoptagtoolresponse

\stepbar{Step 5}
\tagthink

Because the intended visual route is not grounded, I fall back to a plausible flag color. I confuse the red field of the flag with the diagonal band that the question asks about.

\stoptagthink

\vspace{0.3em}
\taganswer
\begin{traceblock}
Red
\end{traceblock}
\stoptaganswer

\vspace{0.4em}
\textbf{Model Answer:} \textbf{\textcolor{red!75!black}{Red}}\\
\textbf{Correct Answer:} \textbf{\textcolor{green!60!black}{White}}

\stepbar{Expected reasoning}
\begin{itemize}
  \setlength{\itemsep}{2pt}
  \item COP30 itinerary page with branded double-decker buses $\rightarrow$ inspect the bus graphics.
  \item Use the guardian identity on the bus bodies as a visual pivot, rather than treating the vehicles as the endpoint.
  \item Return to the COP30 mascot / guardian route and recover the host city branch.
  \item Move from Belém to the state branch, Pará.
  \item Open the official state-flag route.
  \item Ignore the star and inspect the diagonal band.
  \item The band is white.
\end{itemize}

\stepbar{Why this fails}
\begin{itemize}
  \setlength{\itemsep}{2pt}
  \item The model performs several searches, but they are almost entirely broad textual queries.
  \item It never grounds the bus graphics as a visual pivot.
  \item It therefore fails to recover the COP30 guardian identity and the Pará state branch.
  \item At the final step, it confuses the red field of the flag with the diagonal band.
  \item The failure illustrates a single-chain Level~3 bottleneck: the agent must use an intermediate visual clue to redirect the search path, not merely search for related webpages.
\end{itemize}

\end{trajbox}

%\section{Limitations}
%Although InterLV-Search is designed to evaluate interleaved multimodal search, Level~3 is constructed in the open web, where evidence sources are inherently noisy, redundant, and unevenly structured. We guide the generation process toward fine-grained visual cues and apply filtering to reduce text-only shortcuts, but a small portion of examples may still admit alternative textual paths due to overlapping webpages, rich captions, or duplicated descriptions online. As a result, some Level~3 instances may not require visual evidence as strongly as intended. Future work can further improve open-web construction by introducing stricter visual-dependence verification, richer source provenance checks, and more fine-grained trajectory-level validation.

\end{document}